\documentclass[10pt,journal,compsoc]{IEEEtran}
\usepackage{hyperref}
\usepackage{url}
\usepackage{booktabs}
\usepackage{graphicx}
\usepackage{pgfplots}
\usepackage{amsfonts}
\usepackage{bbold}
\usepackage{algorithmic}
\usepackage{algorithm}
\usepackage{setspace}
\usepackage{multirow}
\usepackage{multicol}
\usepackage{tikz}
\usepackage{subfigure}
\usepackage{arydshln}
\usepackage{textcase}
\usepackage{amsmath}
\usepackage{ragged2e}

\newcommand\tikzmark[2]{%
	\tikz[remember picture,baseline] \node[above, outer sep=0pt, inner sep=0pt] (#1){\phantom{#2}};%
} 
\newcommand\link[2]{%
	\begin{tikzpicture}[remember picture, overlay, >=stealth, shift={(0,0)}]
		\draw[->] (#1) to (#2);
	\end{tikzpicture}%
}
\definecolor{color1}{RGB}{27,158,119}
\definecolor{color3}{RGB}{217,95,2}
\definecolor{color2}{RGB}{117,112,179}
\definecolor{ForestGreen}{RGB}{13, 130, 113} 

\ifCLASSOPTIONcompsoc
  \usepackage[nocompress]{cite}
\else
  \usepackage{cite}
\fi

\ifCLASSINFOpdf
\else
\fi

\begin{document}

\title{Cross-lingual Transferring of Pre-trained Contextualized Language Models}

\author{Zuchao Li, Kevin Parnow, Hai Zhao, Zhuosheng Zhang, \\  Rui Wang, Masao Utiyama, 
	and Eiichiro Sumita
	\IEEEcompsocitemizethanks{
		\IEEEcompsocthanksitem Z. Li, K. Parnow,  H. Zhao, Z. Zhang, and R. Wang are with the Department of Computer Science and Engineering, Shanghai Jiao Tong University, and also with Key Laboratory of Shanghai Education Commission for Intelligent Interaction and Cognitive Engineering, Shanghai Jiao Tong University, and also with MoE Key Lab of Artificial Intelligence, AI Institute, Shanghai Jiao Tong University. \protect\\
		E-mail: \{charlee, parnow, zhangzs\}@sjtu.edu.cn, zhaohai@cs.sjtu.edu.cn, wangrui.nlp@gmail.com. M. Utiyama, and E. Sumita are with National Institute of Information and Communications Technology (NICT), Kyoto, Japan, 619-0289. E-mail: \{mutiyama, eiichiro.sumita\}@nict.go.jp. Tel:+81-0774-6986. \protect\\
		
		The corresponding author is Hai Zhao. Part of this work was finished when Zuchao Li, Zhuosheng Zhang were internship research fellows at NICT, and Rui Wang was with NICT. \protect\\
		
		This paper was partially supported by National Key Research and Development Program of China (No. 2017YFB0304100), Key Projects of National Natural Science Foundation of China (No. U1836222 and No. 61733011), and Huawei long term AI project, Cutting-edge Machine Reading Compression and Language Model. \protect\\
	}
}

\markboth{ARXIV, December~2020}%
{Li \MakeLowercase{\textit{et al.}}: Cross-lingual Transferring of Pre-trained Contextualized Language Models}

\IEEEtitleabstractindextext{%
\begin{abstract}
\justifying
Though the pre-trained contextualized language model (PrLM) has made a significant impact on NLP, training PrLMs in languages other than English can be impractical for two reasons: other languages often lack corpora sufficient for training powerful PrLMs, and because of the commonalities among human languages, computationally expensive PrLM training for different languages is somewhat redundant.
In this work, building upon the recent works connecting cross-lingual model transferring and neural machine translation, we thus propose a novel cross-lingual model transferring framework for PrLMs: \textsc{TreLM}. 
To handle the symbol order and sequence length differences between languages, we propose an intermediate ``TRILayer" structure that learns from these differences and creates a better transfer in our primary translation direction, as well as a new cross-lingual language modeling objective for transfer training. 
Additionally, we showcase an embedding aligning that adversarially adapts a PrLM's non-contextualized embedding space and the TRILayer structure to learn a text transformation network across languages, which addresses the vocabulary difference between languages. 
Experiments on both language understanding and structure parsing tasks show the proposed framework significantly outperforms language models trained from scratch with limited data in both performance and efficiency. 
Moreover, despite an insignificant performance loss compared to pre-training from scratch in resource-rich scenarios, our cross-lingual model transferring framework is significantly more economical.
\end{abstract}

}

\maketitle

\IEEEdisplaynontitleabstractindextext

%
\IEEEpeerreviewmaketitle

\IEEEraisesectionheading{\section{Introduction}\label{sec:introduction}}

\begin{figure*}
	\centering
	\includegraphics[width=1.0\textwidth]{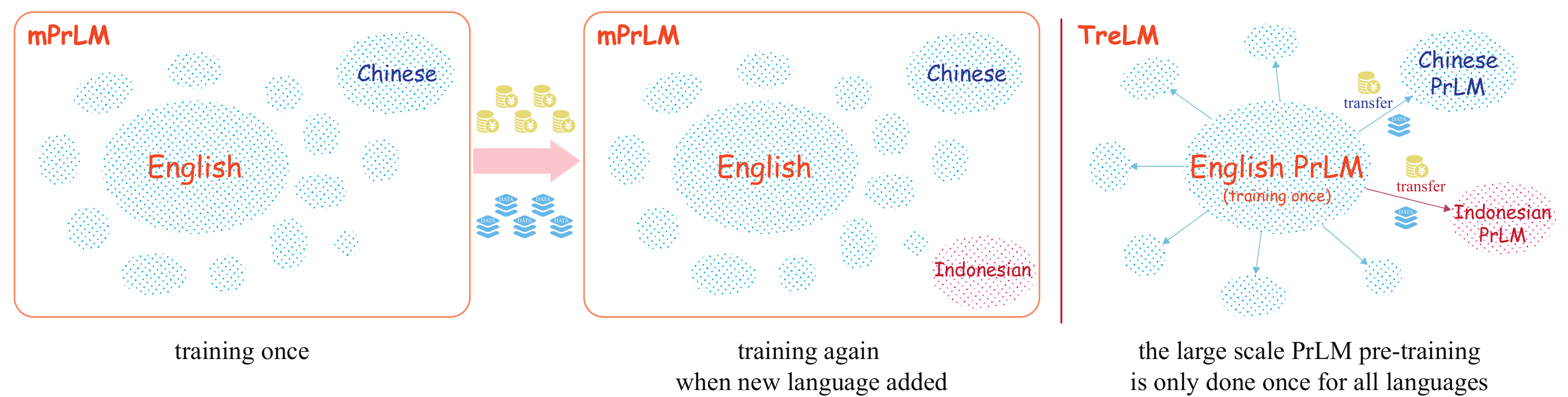}
	\caption{Typical language transfer scenarios for mPrLM and \textsc{TreLM}.}
	\label{fig:scenario}
\end{figure*}

\IEEEPARstart{R}{ecently}, the pre-trained contextualized language model has greatly improved performance in natural language processing tasks and allowed the development of natural language processing to extend beyond the ivory tower of research to more practical scenarios.
Despite their convenience of use, PrLMs currently consume and require increasingly more resources and time. In addition, most of these PrLMs are concentrated in English, which prevents the users of different languages from enjoying the fruits of large PrLMs. Thus, the task of transferring the knowledge of language models from one language to another is an important task for two reasons. First, many languages do not have the data resources that English uses to train such massive and data-dependent models. This causes a disparity in the quality of models available to English users and users of other languages. Second, languages share many commonalities - for efficiency's sake, transferring knowledge between models rather than wasting resources training new ones is preferable. 
\textcolor{black}{Multilingual PrLMs (mPrLMs) also aim to leverage languages' shared commonalities and lessen the amount of language models needed, but they accomplish this by jointly pre-training on multiple languages, which means when they encounter new languages, they need to be pre-trained from scratch again, which causes a waste of resources. This is distinct from using TreLM to adapt models to new languages because TreLM foregoes redoing massive pre-training and instead presents a much more lightweight approach for transferring a PrLM. mPrLMs can risk their multilingualism and finetune on a specific target language, but we will demonstrate that using TreLM to transfer an mPrLM actually leads to better performance than solely finetuning.} 
Therefore, in order to allow more people to benefit from the PrLM, we aim to transfer the knowledge stored in English PrLMs to models for other languages. \textcolor{black}{The differences in training for new languages with mPrLMs and \textsc{TreLM} are shown in Figure \ref{fig:scenario}.}

Machine translation, perhaps the most common cross-lingual task, is the task of automatically converting source text in one language to text in another language; that is, the machine translation model converts the input consisting of a sequence of symbols in some language into a sequence of symbols in another language; i.e., it follows a sequence-to-sequence paradigm. Language has been defined as ``\textit{a sequence that is an enumerated collection of symbols in which repetitions are allowed and order does matter}" \cite{chomsky2002syntactic}. From this definition, we can derive three important differences in the sequences of different languages: \textit{symbol sets}, \textit{symbol order}, and \textit{sequence length}, which can also be seen as three challenges for machine translation and three critical issues that we need to address in migrating a PrLM across languages.

In this work, to resolve these critical differences in language sequences, we propose a novel framework that enables rapid cross-lingual model transferring for PrLMs and reduces loss when only limited monolingual and bilingual data are available. To address the first aforementioned issue, symbol sets, we employ a new shared vocabulary and adversarially align our target embedding space with the raw embedding of the original PrLMs. For the symbol order and sequence length issues, our approach draws inspiration from neural machine translation methods that overcome the differences between languages \cite{bahdanau2014neural}, and we thus propose a new cross-lingual language modeling objective, CdLM, which tasks our model with predicting the tokens for a  from its parallel sentence in the target language. To facilitate this, we also propose a new ``TRILayer" structure, which acts as an intermediary layer that evenly splits our models' encoder layers set into two halves and serves to convert the source representations to the length and order of the target language. Using parallel corpora for a given language pair, we train two models (one in each translation direction) initialized with the desired pre-trained language model's parameters. Combining the first half of our target-to-source model's encoder layer set and the second half of our source-to-target model's encoder layer set, we are thus able to create a full target-to-target language model. During training, we use three separate phases for the proposed framework, where combinations of Masked Language Modeling (MLM), the proposed CdLM, and other secondary language modeling objectives are used.

We conduct extensive experiments on Chinese and Indonesian, as well as German and Japanese,  in challenging situations with limited data and transfer knowledge from English PrLMs. On several natural language understanding and structure parsing tasks, BERT \cite{devlin-etal-2019-bert} and RoBERTa \cite{liu2019roberta} PrLM models that we migrate using our proposed framework improve the performance of downstream tasks compared to monolingual models trained from scratch and models pre-trained in a multilingual setting. Moreover, statistics show that our framework also has advantages in terms of training costs.

\section{Related Work}

Because of neural networks' reliance on heavy amounts of data, model transferring has been an increasingly popular method of exploiting otherwise irrelevant data in recent years. It has seen many applications and has been used particularly often in Machine Translation \cite{zoph2016transfer, dabre-etal-2017-empirical, qi-etal-2018-pre, nguyen-chiang-2017-transfer, gu-etal-2018-meta, kocmi-bojar-2018-trivial, neubig-hu-2018-rapid, kim2019effective, aji-etal-2020-neural}, in which model transferring is generally used to improve the model's translation performance in a low resource scenario using the knowledge of a model trained in a high resource scenario. In addition to cross-lingual situations, model transferring has also been applied to adapt across domains in the POS tagging \cite{schnabel-schutze-2013-towards} and syntactic parsing \cite{mcclosky-etal-2010-automatic, rush-etal-2012-improved} tasks, for example, as well as specifically for adapting language models to downstream tasks \cite{chronopoulou-etal-2019-embarrassingly, houlsby2019parameterefficient}. One particular difference between our method and many model transferring methods is that we do not exactly use the popular "Teacher-Student" framework of model transferring, which is particularly often used in knowledge distillation \cite{hinton2015distilling, sanh2020distilbert} - transferring knowledge from a larger model to a smaller model. We instead use two "student" models, and unlike traditional methods, these student models do not share a target space with their teacher (the language is different), and their parameters are initialized with the teacher's parameters rather than being probabilistically guided by the teacher during training. 

When using model transferring for cross-lingual training, there have been various solutions for the vocabulary mismatch. \cite{zoph2016transfer} did not find vocabulary alignment to be necessary, while \cite{nguyen-chiang-2017-transfer} and \cite{kocmi-bojar-2018-trivial} used joint vocabularies, and \cite{kim2019effective} made use of cross-lingual word embeddings. One particular work that inspired us is that of \cite{lample2018word}, who also used an adversarial approach to align word embeddings without any supervision while achieving competitive performance for the first time. This succeeded the work of  \cite{zhang-etal-2017-adversarial}, who also used an adversarial method but did not achieve the same performance. Also like our aligning method, \cite{xu-etal-2018-unsupervised-cross} took advantage of the similarities in embedding distributions and cross-lingually transferred monolingual word embeddings by simultaneously optimizing based on distributional similarity in the embedding space and the back-translation loss. 

\textcolor{black}{Several works have also explored adapting the knowledge of large contextualized pre-trained language models to more languages, which pose a much more complicated problem compared to transferring non-contextualized word embeddings. 
	The previous mainstream approach for accommodating more languages is using mPrLMs. Implicitly joint multilingual models, such as m-BERT \cite{devlin-etal-2019-bert}, XLM \cite{conneau2019cross}, XLM-R \cite{conneau2019unsupervised}, and mBART \cite{liu2020multilingual}, are usually evaluated on multi-lingual benchmarks such as XTREME \cite{hu2020xtreme} and XGLUE \cite{Liang2020XGLUEAN}, while some works use bilingual dictionaries or sentences for explicit cross-lingual modeling with mPrLMs \cite{schuster-etal-2019-cross,mulcaire-etal-2019-polyglot,liu-etal-2019-investigating,Cao2020Multilingual}. 
	Transferring monolingual PrLMs, another research branch, is relatively new.} \cite{artetxe-etal-2020-cross} presented a monolingual transformer-based masked language model that was competitive with multilingual BERT when transferred to a second language. To facilitate this, they did not rely on a shared vocabulary or joint training (to which multilingual models' performance is often attributed) and instead simply learned a new embedding matrix through MLM in the new language while freezing parameters of all other layers. \cite{tran2020english} used a similar approach, though instead of randomly initialized embeddings, he used a sparse word translation matrix on English embeddings to create word embeddings in the target language, reducing the training cost of the model.  

\section{TreLM}

\textcolor{black}{Cross-lingual Transferring of Language Modeling (\textsc{TreLM}) is a framework that rapidly migrates existing PrLMs.  In this framework, the embedding space of a source language is linearly aligned with that of a target using an adversarial embedding alignment, which we experimentally verified was effective due to shared spatial structure similarities. Leveraging joint learning, we propose a novel pre-training objective, CdLM, and unify it with MLM into one format. In regards to model structure, we proposed TRILayer, an intermediary transfer layer, to support language conversion during the CdLM training process.}

\subsection{Adversarial Embedding Aligning}\label{subsec:adv_emb}

Since the symbol sets in different languages are different, the first step in the cross-lingual migration of PrLMs is to supplement or even replace their vocabularies. In our proposed framework, to make the best use of the commonalities between languages, we choose to use a shared vocabulary with multiple languages rather than replace the original language vocabulary with one for the new language. In addition, in current PrLMs, a subword vocabulary is generally adopted in order to better mitigate out-of-vocabulary (OOV) problems caused by limited vocabulary size. 
To accommodate the introduction of a shared vocabulary, it is necessary to jointly re-train the subword model to ensure that some common words in different languages are consistent in subword segmentation, which leads to the problem that some tokens in the newly acquired subword vocabulary are different from those in the original subword vocabulary, though they belong to the same language.
To address this issue, we consider the most complicated case, in which the vocabulary is completely replaced by a new one.
Consequently, we assume that there are two embedding spaces: one is the embedding of the original vocabulary, which is well-trained in the language pre-training process, and the other is the embedding of the new vocabulary, yet to  be trained.

When considering raw embeddings and non-contextualized embeddings (e.g. Word2vec), it is easy to see their training objectives are similar in theory. The only differences are the addition of context and the change in model structure to accommodate language prediction. Despite these differences, non-contextualized embeddings can be used to simulate the raw embeddings in a  PrLM that we aim to replace (refer to Appendix \ref{subsec:non_emb_simulation} for a detailed explanation). Although the two embedding spaces we consider are similar in structure, they may be at different positions in the whole real embedding space, so an extra alignment process is required, and although common tokens may exist, due to the inconsistent token granularity from using byte-level byte-pair encoding (BBPE) \cite{radford2019language}, a matching token of the two embedding spaces cannot be utilized for embedding space alignment, as it is likely to represent different meanings. Therefore, inspired by \cite{lample2018word}, we present an adversarial approach for aligning the word2vec embedding space to the PrLM's raw embedding space without supervision. With this approach, we aim to minimize the differences between the two embedding spaces brought about by different similarity forms.

We define $\mathcal{U} = \{u_1, u_2, ..., u_m\}$ and $\mathcal{V} = \{v_1, v_2, ..., v_n\}$ as the two embedding spaces of $m$ and $n$ tokens from the PrLM and word2vec training, respectively. In the adversarial training approach,  a linear mapping $W$ is trained to make the spaces $W\mathcal{V} = \{Wv_1, Wv_2, ..., Wv_n\}$ and $\mathcal{U}$ close as possible, while a discriminator $D$ is employed to discriminate between tokens randomly sampled from spaces $W\mathcal{V}$ and $\mathcal{U}$. Let $\theta_{adv}$ denote the parameters of the adversarial training model and the probabilities $P_{\theta_{adv}}(\mathbb{1}(z)|z)$ and $P_{\theta_{adv}}(\mathbb{0}(z)|z)$ indicate whether or not the sampling source prediction is the same as its real space for a vector $z$. Therefore, the discrimination training loss $\mathcal{L}_D(\theta_D|W)$ and the mapping training loss $\mathcal{L}_D(W|\theta_D)$ are defined as:
\begin{equation}
	\begin{aligned}
	\mathcal{L}_D(\theta_D|W) = -\frac{1}{n} \sum_{i=1}^{n} \log P_{\theta_{adv}}(\mathbb{1}(Wv_i)|Wv_i) \\
	- \frac{1}{m} \sum_{i=0}^{m} \log P_{\theta_{adv}}(\mathbb{1}(u_i)|u_i),
	\end{aligned}
\end{equation}
\begin{equation}
	\begin{aligned}
	\mathcal{L}_W(W|\theta_D) = -\frac{1}{n} \sum_{i=1}^{n} \log P_{\theta_{adv}}(\mathbb{0}(Wv_i)|Wv_i) \\
	- \frac{1}{m} \sum_{i=0}^{m} \log P_{\theta_{adv}}(\mathbb{0}(u_i)|u_i),
	\end{aligned}
\end{equation}
where $\theta_D$ are the parameters of discriminator $D$, which  is implemented as a multilayer perceptron (MLP) with two hidden layers and Leaky-ReLU as the activation function.

During the adversarial training, the discriminator parameters $\theta_D$ and $W$ are optimized successively with discrimination training loss and mapping training loss. To enhance the effect of embedding space alignment, we adopted the same techniques of iterative refinement and cross-domain similarity local scaling as \cite{lample2018word} did. While the two embedding spaces in \cite{lample2018word} both can be updated by gradient, we consider $\mathcal{U}$ as the goal spatial structure and hence fix $\mathcal{U}$ throughout the training process, and we update $\mathcal{W}$ to better align $\mathcal{V}$.

\begin{figure}[h]
	\centering
	\includegraphics[width=0.5\textwidth]{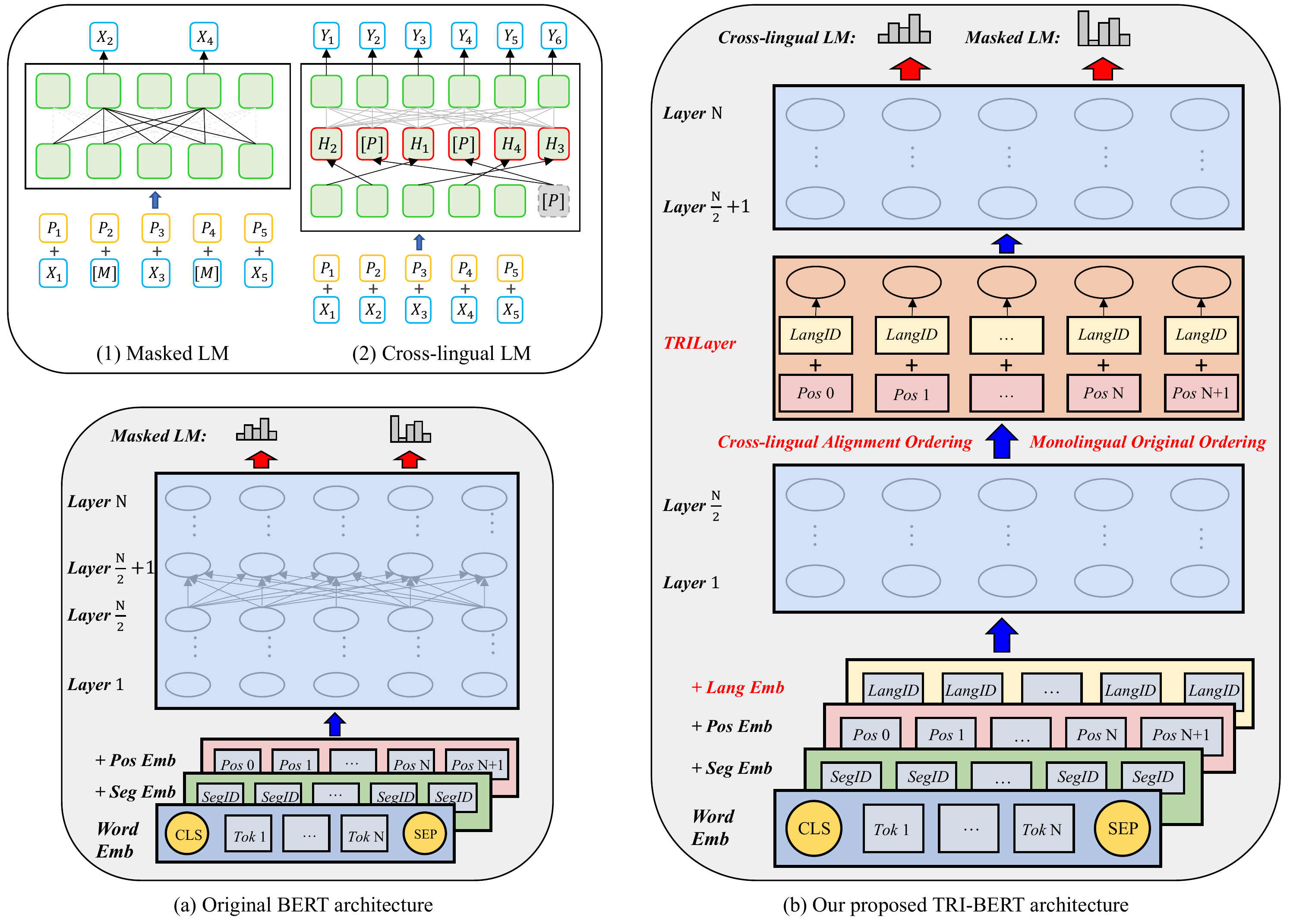}
	\caption{The proposed model architecture,  taking BERT as an example.}
	\label{fig:model}
\end{figure}

\subsection{TRILayer and CdLM}\label{subsec:cdlm}

For the disparities in symbol sets of different languages and different pre-trained models, we employ embedding space alignment, while for the issues of the symbol order and sequence length, unlike previous work, we do not assume that the model can implicitly learn these differences, and we instead leverage language embeddings and explicit alignment information and propose a novel Cross-Lingual Language Modeling (CdLM) training objective  and a Transferring Intermediate Layer (TRILayer) structure as a pivot layer in the model to bridge the differences of the two languages. To clearly explain our training approach, we take the popular PrLM BERT as a basis for introduction.

In the original BERT (as shown in Figure\ref{fig:model}(a)), Transformer \cite{vaswani2017attention} is taken as the backbone of model, which takes tokens and their positions in a sequence as input before encoding this sequence into a contextualized representation using multiple stacked multi-head self-attention layers. 
During the pre-training process, BERT predominantly adopts an MLM training objective, in which a {\tt [MASK]} (also written as {\tt [M]}) token is used to replace a token in the sequence selected by a predetermined probability, and the original token is predicted as the gold target. Formally speaking, given a sentence $X = \{x_1, x_2, ..., x_T\}$ and $\mathcal{M}$, the set of masked positions, the training loss $\mathcal{L}_{\textit{MLM}}$ for the MLM objective is:
$$\mathcal{L}_{\textit{MLM}}(\theta_{\textit{LM}}) = -\sum_{i=1}^{|\mathcal{M}|} \log P_{\theta_{\textit{LM}}}(x_{\mathcal{M}_i}|X_{\backslash\mathcal{M}}),$$
where $\theta_{\textit{LM}}$ are the parameters of BERT, $|\mathcal{M}|$ is the length of set $\mathcal{M}$, and $X_{\backslash\mathcal{M}}$ indicates the sequence after masking. An example of MLM training is shown in the top-left region of Figure \ref{fig:model}.

Much work in the field of machine translation suggests that the best way to model transferring across languages is through translation learning because the machine translation model must address all three of the above-described language differences in the training process. Therefore, we take inspiration from the design of machine translation, especially the design of non-autoregressive machine translation, and propose a Cross-Lingual Language Modeling (CdLM) objective. CdLM is just like a traditional language modeling objective, except across languages, so given an input of source tokens, it generates tokens in a separate target language.  With this proposed objective, we aim to make as few changes as possible to the existing PrLM and thus introduce a Translation/Transfer Intermediate Layer (``TRILayer") structure, which bridges two opposing half-models to create our final model. 

First, in the modified version of BERT for cross-lingual model transferring, we add a language embedding $E_{\textit{lng}}$ following the practice of \cite{conneau2019cross} to indicate the current language being processed by the model. This is important because the model will handle both the source and target languages simultaneously in 2 of our 3 training phases (described in next subsection). The new input embedding is:
$$E_{inp} = E_{wrd} + E_{seg} +E_{pos} + E_{lng},$$
where $E_{wrd}$, $E_{seg}$, and $E_{pos}$ are the word (token) embedding, segment embedding, and position embedding, respectively.

Next, we denote $N$ as the number of stacked Transformer layers ($L = \{l_1, l_2, ..., l_N\}$) in BERT and split the BERT layers into two halves $L_{\leq\frac{N}{2}} = \{1_1, ..., l_{\frac{N}{2}}\}$ and $L_{>\frac{N}{2}} = \{l_{\frac{N}{2}+1}, l_{\frac{N}{2}+2}, ...,l_N\}$. The TRILayer is placed between the two halves (making the total number of layers $N+1$) and functions as a pivot. In the $L_{\leq\frac{N}{2}}$ half, the input embedding is encoded by its Transformer layers to hidden states 
$H_i = \textsc{Transformer}_i(H_{i-1}),$
in which $H_0 = E_{inp}$ and $\textsc{Transformer}_i$ indicates the $i$-th Transformer layer in the model. 

Before the outputs of the $L_{\leq\frac{N}{2}}$ half are fed into the TRILayer,  the source hidden representation $H_{\frac{N}{2}}$ is reordered according to new order $O$.
During CdLM training, for source language sentence $X = \{x_1, x_2, ..., x_T\}$, a possible translation sentence $Y = \{y_1, y_2, ..., y_{T^{'}}\}$ is provided. To find the new order, explicit alignment information between the transfer source and target sentences is obtained using an unsupervised external aligner tool. We define the source-to-target alignment pair set as:
\begin{equation}
	\begin{aligned}
		\mathcal{A}_{X \rightarrow Y} &= \textnormal{ALIGN}(X, Y) = \{(x_{\textsc{AlnIdx}(y_1)}, y_1), \\
		&(x_\textsc{AlnIdx}(y_2), y_2), ...,(x_\textsc{AlnIdx}(y_{T^{'}}), y_{T^{'}})\},
	\end{aligned}
\end{equation}
where $\textsc{AlnIdx}(\cdot)$ is a function that returns the alignment index in the source language or $x_{\textnormal{null}}$ when there is no explicit alignment between the token in the target language and any source language token. $x_{\textnormal{null}}$ represents a special placeholder token {\tt [P]} that is always appended to the inputs. 
Finally, the source hidden representation $H_{\frac{N}{2}}$ is reordered according to the new order $O = \{\textsc{AlnIdx}(y_1), \textsc{AlnIdx}(y_2), ..., \textsc{AlnIdx}(y_{T^{'}})\}$ from alignment set $\mathcal{A}_{X \rightarrow Y}$, creating $H_{\frac{N}{2}}^O$. 

Thus, the resultant hidden representation $H_{\frac{N}{2}}^O$ is in the order of the target language and is consistent with the target sequence in length, making it usable for language modeling prediction. 
Unfortunately, the position information is lost in reordering. To combat this, the position embedding and language embedding will be reintegrated as follows:
$$H_{\textnormal{TL}} = \textsc{Transformer}_{\textnormal{TL}}(H_{\frac{N}{2}}^O + E_{lng^{Y}} + E_{pos} ),$$
where $H_{\textnormal{TL}}$ is the output of TRILayer, $\textsc{Transformer}_{\textnormal{TL}}$ is the Transformer structure inside the TRILayer, and $\textbf{E}_{lng^{Y}}$ is the target language embedding. Next, the $H_{\textnormal{TL}}$ is encoded in the $L_{>\frac{2}{N}}$ half as done for the $L_{\leq\frac{N}{2}}$ half (let $H_{\frac{N}{2}} = H_{\textnormal{TL}}$ for the $L_{>\frac{N}{2}}$ half) to predict the final full sequence of the target language. The model is trained to minimize the loss $\mathcal{L}_{\textit{CdLM}}$, which is:
$$\mathcal{L}_{\textit{CdLM}}(\theta_{\textit{LM}}) = -\sum_{i=1}^{T^{'}} \log P_{\theta_{\textit{LM}}}(y_i|X, \mathcal{A}_{X \rightarrow Y}).$$

To enable MLM and CdLM to train models simultaneously rather than through successive optimization, we provide a unified view for MLM and CdLM language modeling:
$$\mathcal{L}_{\textit{ULM}}(\theta_{\textit{LM}}) = -\sum_{i=1}^{T_{\textnormal{max}}} \mathbb{1}(i \in C)  \log P_{\theta_{\textit{LM}}}(w_i|S, \mathcal{A}),$$
where $T_{\textnormal{max}}$ denotes the maximum sequence length for language modeling, $S$ is the input sequence, $w_i$ is the $i$-th token in output sequence $W$, $C$ is the set of positions to be predicted, and $A$ is the alignment between the input and output sequence. 
Both the input and output sequences are padded to the maximum sequence length $T_{\textnormal{max}}$ during training.
$\mathbb{1}(i \in C)$ represents the indicator function and equals 1 when $i$-th position exists in the set for the parse to be predicted and 0 otherwise. In MLM, $S = X_{\backslash C}$ , $\mathcal{A} = \{(1, 1), (2, 2), ..., (T_{\textnormal{max}} , T_{\textnormal{max}})\}$ is a successive alignment, and $W = X$, while in CdLM, $S = X$, $\mathcal{A} = \mathcal{A}_{X \rightarrow Y}$, and $W = Y$. Due to the unified language modeling abstractions of MLM and CdLM, the input and output forms, as well as the internal logic of their models, are the same. Therefore, models can be trained with the two objectives in the same mini-batch, which enhances the stability of transfer training.

\noindent\textbf{Difference between MLM, TLM, BRLM, and CdLM} As stated in the original MLM objective, the model can only learn from monolingual data. Though a joint MLM training can be performed across languages, there is still a lack of explicit language cues for guiding the model in distinguishing language differences. \cite{conneau2019cross} proposed a Translation Language Modeling (TLM) objective as an extension of the MLM objective. The TLM objective leverages bilingual parallel sentences by concatenating them into single sequences as in the original BERT and predicts the tokens masked in the concatenated sequence.  This encourages the model to predict the masked part in a bilingual context. \cite{ji2020cross} further proposed a BRidge Language Modeling (BRLM) built on the TLM, benefiting from explicit alignment information or additional attention layers that encourage word representation alignment across different languages. These MLM variants drive models to learn explicit or implicit token alignment information across languages and have been shown effective in machine translation compared to the original MLM, but for the cross-lingual model transferring of PrLMs, modeling the order difference and semantic equivalence in different languages is still not enough. Since both contexts in MLM variants have been exposed to the model, whether the prediction of the masked part depends on the cross-lingual context or the context of its own language is unknown, as it lacks explicit clues for cross-lingual training. In our proposed CdLM, we use sentence alignment information for explicit ordering. The model is exposed to both the transfer source and transfer target languages at the same time, during which the input is a sequence of the source language, and the prediction goal is a sequence of the target language. Thus, we convert translation into a cross-language modeling objective,  which gives a clear supervision signal for cross-lingual transferring.

\subsection{Triple-phase Training}
In our \textsc{TreLM} framework, 
the whole training process is divided into three phases with different purposes but the same design goal: minimize the number of parameter updates as much as possible to speed up convergence and enhance training stability. The three phases are \textit{commonality training}, \textit{transfer training}, and \textit{language-specific training}. In the \textit{commonality learning} phase, only the target language MLM objective is used, while in the \textit{transfer training} phase, CdLM and target language MLM objectives are both used at the same time, and in the final \textit{language-specific learning} phase, target language MLM and other secondary language modeling objectives are adopted.

\noindent\textbf{\textit{Commonality Training}} Though languages are very different on the surface, they also share a lot of underlying commonalities, often called linguistic universals or cross-linguistic generalizations. We therefore take advantage of these commonalities between languages and jointly learn the transferring source and target languages. In this phase, the parameters of the position, segment embedding, and Transformer layers are initialized with original BERT, the TRILayer is initialized with parameters of Transformer layer $L_{\frac{N}{2}}$, the word embedding is initialized with the output of the adversarial embedding aligning, and orthogonal weight initializations are adopted for the language embedding. For this phase, the model is trained by joint MLM with monolingual inputs from both the source and target languages. Moreover, in this training process, to make convergence fast and stable, the parameters of BERT's backbone (Transformer) layers are fixed; only the embeddings and TRILayer are updated by the gradient-based optimization based on the joint MLM loss. The final model obtained in this phase is denoted as $\theta_{\textnormal{LM}}^{ct}$.

\noindent\textbf{\textit{Transfer Training}} Since the model is not pre-trained from scratch, making the model aware of changes in inputs is a critical factor for a maximally rapid and accurate migration in the case of limited data. Since there is not enough monolingual data in the target language to allow the model to adapt to the new language, we use the supervisory signal from the two languages' differences and leverage parallel corpora to directly train the model. Specifically, we split the original BERT transformer layers into two halves. With a parallel corpus from the source language to the target language and one from the target language to the source language, we train two corresponding models, both of which are initialized using the parameters learned in the previous phase. In the source-to-target model, only the upper half of the encoder layers is trained, and the lower half is kept fixed, while the converse is true for the target-to-source model. TRILayer then provides cross-lingual order and length adjustment, which is similar to the behavior of a neural machine translation model. Thus, we create two reciprocal models: one whose upper half can handle the target language, and one whose lower half can handle it, which we connect via the TRILayer. Finally, the two trained models are combined as $\theta_{\textnormal{LM}}^{tt}$. We describe the full procedure in Algorithm \ref{alg:tlplm}. 

\begin{algorithm}[H]
	\setstretch{1.25}
	\small
	\centering
	\caption{\small Transfer Training of Pre-trained Contextualized Language Models}\label{alg:tlplm}
	\begin{algorithmic}[1]
		\REQUIRE The commonality pre-trained model parameters $\theta_{\textnormal{LM}}^{ct}$, Languages $\mathfrak{L} = \{lng^X, lng^Y\}$, Parallel training set $\mathfrak{P} = \{(X^{\mathfrak{L}_0}_i, X^{\mathfrak{L}_1}_i)\}_{i=1}^{|\mathfrak{P}|}$, Number of training steps $\mathcal{K}$ \\
		\FOR{$j$ in 0, 1}
		\STATE Initialize model parameters $\theta_{\textnormal{LM}}^{\mathfrak{L}_j \rightarrow \mathfrak{L}_{(1-j)}} \leftarrow \theta_{\textnormal{LM}}^{ct}$
		\IF{$j == 0$} 
		\STATE Fix the parameters of $L_{\leq\frac{N}{2}}$ half of $\theta_{\textnormal{LM}}^{\mathfrak{L}_j \rightarrow \mathfrak{L}_{(1-j)}}$
		\ELSE
		\STATE Fix the parameters of $L_{>\frac{N}{2}}$ half of $\theta_{\textnormal{LM}}^{\mathfrak{L}_j \rightarrow \mathfrak{L}_{(1-j)}}$
		\ENDIF
		\FOR{step in 1, 2, 3, ..., $\mathcal{K}$}
		\STATE Sample batch $(X^{\mathfrak{L}_j}, X^{\mathfrak{L}_{(1-j)}})$ from $\mathfrak{P}$.
		\STATE Alignment information $\mathcal{A}$:  $\mathcal{A}_{\mathfrak{L}_j \rightarrow \mathfrak{L}_{(1-j)}} \leftarrow \textnormal{ALIGN}(X^{\mathfrak{L}_j}, X^{\mathfrak{L}_{(1-j)}})$
		\STATE CdLM Loss: $\mathcal{L}_{\textnormal{CdLM}} \leftarrow - \sum \log P_{\theta_{\textnormal{LM}}^{\mathfrak{L}_j \rightarrow \mathfrak{L}_{(1-j)}}}(X^{\mathfrak{L}_{(1-j)}}|X^{\mathfrak{L}_j}, \mathcal{A}_{\mathfrak{L}_j \rightarrow \mathfrak{L}_{(1-j)}})$
		\STATE Masked version of $X^{\mathfrak{L}_1}$: $X^{\mathfrak{L}_1}_{\backslash M} \leftarrow \textnormal{MASK}(X^{\mathfrak{L}_1})$
		\STATE MLM Loss: $\mathcal{L}_{\textnormal{MLM}} \leftarrow - \sum \log P_{\theta_{\textnormal{LM}}^{\mathfrak{L}_j \rightarrow \mathfrak{L}_{(1-j)}}}(X^{\mathfrak{L}_1}_{M}|X^{\mathfrak{L}_1}_{\backslash M})$
		\STATE CdLM+MLM Update: $\theta_{\textnormal{LM}}^{\mathfrak{L}_j \rightarrow \mathfrak{L}_{(1-j)}} \leftarrow \textnormal{optimizer\_update}(\theta_{\textnormal{LM}}^{\mathfrak{L}_j \rightarrow \mathfrak{L}_{(1-j)}}, \mathcal{L}_{\textnormal{CdLM}}, \mathcal{L}_{\textnormal{MLM}})$
		\ENDFOR
		\ENDFOR
		\STATE Combine the two obtained models as $\theta_{\textnormal{LM}}^{tt}$ by choosing the $L_{>\frac{N}{2}}$ half model parameters from model $\theta_{\textnormal{LM}}^{\mathfrak{L}_0 \rightarrow \mathfrak{L}_1}$ and $L_{\leq\frac{N}{2}}$ half model parameters from model $\theta_{\textnormal{LM}}^{\mathfrak{L}_1 \rightarrow \mathfrak{L}_0}$ and average the other parameters (such as embedding and TRILayer parameters) of the two models
		\ENSURE Learned model $\theta_{\textnormal{LM}}^{tt}$
	\end{algorithmic}
\end{algorithm}

\noindent\textbf{\textit{Language-specific Training}} During the language-specific training phase, we only use the monolingual corpus of the target language and further strengthen the target language features for the model obtained in the transfer training phase. We accomplish this by using the MLM objective and other secondary objectives such as Next Sentence Prediction (NSP).

\section{Experiments}

In this section, we discuss the details of the experiments undertaken for this work. We conduct experiments based on English PrLMs\footnote{Our code is available at \url{https://github.com/agcbi2017/TreLM}.}. We transfer via English-to-Chinese and English-to-Indonesian directions for the purpose of comparing with previous recent work. 
From English to Chinese and English to Indonesian, we transfer two pre-trained contextualized language models: BERT and RoBERTa. 
Our performance evaluation on the migrated models is mainly conducted on two types of downstream tasks: language understanding and language structure parsing. \textcolor{black}{Please refer to Appendix \ref{subsec:tasks} for introductions of tasks and baselines.}
\textcolor{black}{We note that the comparisons between models trained using \textsc{TreLM} and the monolingual and multilingual PrLMs trained from scratch on the target language (see Table \ref{table:clue}) is only for illustrating the relative performance loss of the model produced by \textsc{TreLM}. These models are not directly comparable, as we intentionally use less data to train models when using \textsc{TreLM}. Continuing to pre-train the PrLMs on the target language would also obviously further improve their performance, but this is not our main focus.}

\begin{table*}[!htb]
	\small
	\centering
	\setlength{\tabcolsep}{3pt}
	\caption{Results on the CLUE development datasets.}\label{table:clue}
	\begin{tabular}{ l  c c c cc c c cc c c}
		\toprule
		{\bf Models} & \multicolumn{3}{c}{{\bf Sentence-Pair}} & & \multicolumn{3}{c}{{\bf Single-Sentence}} & & \multicolumn{3}{c}{{\bf MRC}} \\
		\cmidrule{2-4} \cmidrule{6-8} \cmidrule{10-12} & AFQMC& CMNLI  & CSL & & TNEWS  &  IFLYTEK&  WSC  & &  CMRC18 &  CHID & C$^3$	\\
		& (acc) & (acc) & (acc) & & (acc) & (acc) & (acc) & & (EM) & (acc) & (acc) \\
		\midrule
		\multicolumn{12}{l}{\textit{Single-task single models on dev}}\\
		BERT-base	 & 74.16 & 79.47 & 79.63  & & 56.09 & 60.37 & 63.48 & & 64.77 & 82.20 & 65.70 	\\
		m-BERT-base & 70.29 & 79.03 & 79.26 & & 53.71 & 56.63 & 62.82 & & 63.93 & 80.00 & 63.81 \\
		\midrule
		BERT-small & 69.71 & 66.54 & 69.73 &  & 53.22 & 45.40 & 53.29 & &  50.23 & 68.55  &  56.84  \\
		{\bf TRI-BERT-base}	 & 72.98 & 79.44 & 79.34 & & 55.45 & 58.36 & 63.00 & & 63.96 & 80.94 & 65.06 	\\	
		{\bf TRI-BERT-large}	 & 73.41 & 80.50 & 80.59 & & 56.20 & 60.99 & 64.76 & & 66.35 & 82.61 & 66.08  	\\	
		{\bf TRI-RoBERTa-base} & 73.51 & 80.47 & 80.26 & & 55.98 & 61.65 & 63.92 & & 65.76 &  82.02 & 65.98 \\	
		{\bf TRI-RoBERTa-large} & 74.55 & 81.68 & 81.35 &  & 57.02 & 62.24 & 65.16 & & 67.29 & 83.53 & 66.79  	\\	
		\bottomrule
	\end{tabular}   
\end{table*}

\subsection{Training Details}\label{subsec:training_details}
The initial weights for the migration are {\tt BERT-base-cased}, {\tt BERT-large-cased}, {\tt RoBERTa-base}, and {\tt RoBERTa-large}, which are taken from their official sources. 
We use English Wikipedia, Chinese Wikipedia, Chinese News, and Indonesian CommonCrawl Corpora for the monolingual pre-training data.
For all models migrated in the same direction, regardless of their original vocabulary, we used the same single vocabulary that we trained on the joint language data using the WordPiece Subword scheme \cite{schuster2012japanese}. In English-to-Chinese, the vocabulary size is set to 80K and the alphabet size is limited to 30K, while in English-to-Indonesian, the vocabulary size is set to 50K, and the alphabet size is limited to 1K.
With the WordPiece vocabulary, we tokenized the monolingual corpus to train the non-contextualized word2vec embedding of subwords. Using the {\tt fastText} \cite{bojanowski2017enriching} tool and skipgram representation mode, three embedding sizes 128, 768, and 1024 were trained to be compatible the respective pre-trained language models.

In the \textit{commonality training} phase, we sampled 1M sentences of English Wikipedia and either 1M sentences of Chinese Wikipedia or 1M sentences of Indonesian CommonCrawl for the English-to-Chinese and English-to-Indonesian models. We trained the model with 20K update steps with total batch size 128 and set the peak learning rate to 3e-5.

For the \textit{transfer training} phase, we sampled 1M parallel sentences from the UN Corpus \cite{ziemski-etal-2016-united} for English-to-Chinese and 1M parallel sentences from OpenSubtitles Corpus \cite{lison-tiedemann-2016-opensubtitles2016} for English-to-Indonesian. We use the {\tt fastalign} toolkit \cite{dyer-etal-2013-simple} to extract the tokenized subword alignments for CdLM. The two half models are optimized over 20K update steps, and the batch size and peak learning rate are set to 128 and 3e-5, respectively.

In the final phase, \textit{language-specific training}, 2M Chinese and Indonesian sentences were sampled to update their respective models, training for 80K steps with total batch size 128 and initial learning rate 2e-5. In all the above training phases, the maximum sequence length was set to 512, weight decay was 0.01, and we used Adam \cite{kingma2014adam} with $\beta_1 = 0.9$, $\beta_2 = 0.999$.

In addition to our migrated pre-trained models, we also pre-trained a {\tt BERT-small}\footnote{The performance of {\tt BERT-base}  for pre-training from scratch with this limited data is inferior to that of {\tt BERT-small}, so we do not compare it with our migrated models.} model from scratch with data of the same size as our migration process to compare the performance differences between migration and scratch training. For the {\tt BERT-small} model, we started with the  {\tt BERT-base}  hyper-parameters and vocabulary but shortened the maximum sequence length from 512 to 128, reduced the model’s hidden and token embedding dimension size from 768 to 256, set the batch size to 256, and extended the training steps to 240K.

\textcolor{black}{Our TRI-BERT -* and TRI-RoBERTa-* all used the same amount of training data (2M target language monolingual sentences, 1M source language monolingual sentences, and 1M parallel sentences). BERT-small pre-trained from scratch on only the target language, using 5M target language sentences to ensure the training data amount was the same. Compared with the original model, the TRI-* model only has an extra TRI-layer added and some changes in the embedding layer. BERT-base-chinese and m-BERT-base models were downloaded from the official repository, which trained with 25M sentence (much more than our 5M sentences) and more training steps.}

\subsection{Empirical Analysis }

\noindent\textbf{Language Understanding} We first compare the PrLMs transferred by \textsc{TreLM} alongside the results the existing monolingual pre-trained {\tt BERT-base-chinese} and the multilingual pre-trained {\tt BERT-base-multilingual} in Table \ref{table:clue} using the CLUE benchmark.

When comparing with the same model architecture, taking BERT as an example, our model TRI-BERT-base exceeds m-BERT-base and BERT-small and is slightly weaker than original BERT-base. Compared with BERT-small, which is trained from scratch for a longer time, our TRI-BERT-base generally achieves better results on these NLU tasks. This demonstrates that because of the commonalities of languages, models for languages with relatively few resources can benefit from language models pre-trained on languages with richer resources, which confirms our cross-lingual model transferring framework's effectiveness.  

m-BERT is another potential language model migration scheme and has the advantage of supporting multiple languages at the same time; however, in order to be compatible with multiple languages, the unique characteristics of each language are neglected. Our TRI-BERT, which is built on top of BERT-base, instead focuses on and highlights language differences during the model transferring process, which leads to an increase in performance compared to m-BERT.
When TRI-BERT and TRI-RoBERTa have the same model size, TRI-RoBERTa outperforms TRI-BERT, which is consistent with the performance differences between the original RoBERTa and BERT, indicating that our migration approach maintains the performance advantages of PrLMs. 

\begin{table*}
	\centering
	\caption{Dependency parsing results on the Chinese PTB 5.1, CoNLL-2009 Chinese, and Universal Dependency Indonesian GSD test sets. ``$*$" indicates that the result was from our own experiments on the UD dataset based on \cite{dozat2016deep}'s model, and ``$\dag$" indicates that the official BERT paper did not provide Indonesian BERT-base, so we used IndoBERT-base pre-trained by \cite{wilie2020indonlu}.}
	\label{table:dp}
	\begin{tabular}{lcccccccc} 
		\toprule
		\multirow{2}{*}{\bf Models} & \multicolumn{2}{c}{\bf CTB 5.1} & & \multicolumn{2}{c}{\bf CoNLL-09 ZH} & & \multicolumn{2}{c}{\bf UD 2.3 ID GSD} \\ 
		\cmidrule{2-3} \cmidrule{5-6} \cmidrule{8-9} & UAS & LAS & & UAS & LAS & & UAS & LAS \\
		\midrule
		(Dozat and Manning, 2016) \cite{dozat2016deep} & 89.30 & 88.23 & & 88.90 & 85.38 & & 85.93$^*$  & 78.21$^*$  \\ 
		BERT-base & 91.48 & 89.24 & & 92.63 & 89.59 & & 86.69$^\dag$ & 77.97$^\dag$ \\
		m-BERT-base & 89.84 & 87.33 & & 90.98 & 87.70 & & 87.19$\  $ & 79.10$\  $ \\
		\hdashline
		\bf TRI-BERT-base & 89.96 & 87.43 & & 90.94 & 87.81 & & 87.56$\  $ & 79.44$\  $ \\
		\bf TRI-RoBERTa-base & 90.30 & 87.82 & & 91.62 & 88.29 & & 88.42$\  $ & 79.95$\  $ \\
		\midrule
		EN-ID-ZH XLM-MLM & 89.25 & 86.98 & & 90.00 & 86.98 & & 86.64$\  $ & 77.74$\  $ \\
		EN-ID-ZH XLM-MLM+TLM & 89.58 & 87.16 & & 90.53 & 87.37 & & 86.96$\  $ & 78.01$\  $ \\
		\hdashline
		\bf TRI-XLM-en-2048 & 90.66 & 88.20 & & 91.95 & 88.61 & & 88.67$\  $ & 80.30$\  $ \\
		\bottomrule
	\end{tabular}
\end{table*}

\noindent\textbf{Language Structure Parsing}  We report results on dependency parsing for Chinese and Indonesian in Table \ref{table:dp}. As shown in the results, the baseline model has been greatly improved for the PrLM. In Chinese, the performance of BERT-base is far superior to m-BERT-base, which highlights the importance of the unique nature of the language for downstream tasks, especially for refined structural analysis tasks. In Indonesian, IndoBERT \cite{wilie2020indonlu} performs worse than m-BERT, which we suspect is due to IndoBERT's insufficient pre-training. We also compare TRI-BERT-base and IndoBERT-base on Indonesian, whose ready-to-use language resources are relatively small compared to English. We find that although pre-training PrLMs on the available corpora is possible, because of the size of language resources, engineering implementation, etc., our migrated model is more effective than the model pre-trained from scratch. This shows that migrating from the ready-made language models produced from large-scale language training and extensively validated by the community is more effective than pre-training on relatively small and limited language resources. In addition, we also conduct experiments for these pre-trained and migrated models on Chinese SRL. 

\textcolor{black}{mPrLMs are another important and competitive approach that can adapt to  cross-lingual PrLM applications, so we also include several mPrLMs in our comparison on dependency parsing. Specifically, we used XLM, a monolingual and multilingual PrLM pre-training framework, as our basis. For \textsc{TreLM}, we used {\tt XLM-en-2048}, officially provided by \cite{conneau2019cross}, as the source model. The data amount used and the number of training steps are consistent with TRI-BERT/TRI-RoBERTa. In mPrLM, we combined EN, ID, and ZH sentences (including monolingual and parallel sentences) together (10M sentences in total) to train an EN-ID-ZH mPrLM with MLM and TLM objectives. The performance comparison of these three PrLMs on the dependency parsing task is shown in the lower part of Table \ref{table:dp}.}

\textcolor{black}{From the results, we see mPrLMs pre-trained from scratch have no special performance advantage over \textsc{TreLM} when corpus size is constant, and especially when not using the cross-lingual model transferring objective TLM, which models parallel sentences. In fact, our TRI-XLM-en-2048 solidly outperforms its two multilingual XLM counterparts. Monolingual PrLMs generally outperform mPrLMs, which likely leads to the performance advantages shown with monolingual migration. Additionally, like our \textsc{TreLM}, mPrLMs can also finetune on only the target language to improve performance, and leveraging \textsc{TreLM} to transfer an mPrLM leads to even further gains, as seen in Table \ref{table:related_work}.} 

\textcolor{black}{While the two approaches can compete with each other, they have their own advantages in general. In particular, \textsc{TreLM} is more suitable for transferring additional languages that were not considered in the initial pre-training phase and for low-resource scenarios, while mPrLMs have the advantage of being able to train and adapt to multiple languages at once.}

In Table \ref{table:srl}, we compared a model migrated without CdLM to the full one. To compensate for the removal of CdLM, we added a monolingual corpus with the same size as the parallel corpora and trained the model with an extra 80K steps, but despite using more target monolingual data and training steps, the performance was still much better when CdLM  was included.

\begin{table}
	\centering
	\caption{Dependency SRL results on the CoNLL-2009 Chinese benchmark.}\label{table:srl}
	\begin{tabular}{lccc} 
		\toprule
		\multirow{2}{*}{\bf Models} & \multicolumn{3}{c}{\bf CoNLL-09}\\
		\cmidrule{2-4} & P & R & F$_1$\\ 
		\midrule
		(Cai et al., 2018) \cite{cai-etal-2018-full} & 84.7 & 84.0 &  84.3 \\
		+BERT-base  & 86.86 & 87.48 & 87.17 \\
		+m-BERT-base & 85.17 & 85.53 & 85.34 \\
		\midrule
		\bf+TRI-BERT-base & 86.15 & 85.58 & 85.86 \\
		\bf+TRI-RoBERTa-base & 87.08	& 86.99 & 87.03 \\
		\hdashline
		\bf+TRI-RoBERTa-base & \multirow{2}{*}{85.77} & \multirow{2}{*}{85.62} & \multirow{2}{*}{85.69} \\
		\bf \quad (w/o CdLM) & & & \\
		\bottomrule
	\end{tabular}
\end{table}

\begin{figure}
	\centering
	\resizebox{6.8cm}{4.2cm}{
		\pgfplotsset{compat=newest}
		\begin{tikzpicture}[every axis/.append style={font=\large}]
			
			\begin{axis}[
				mark=none, 
				xmax=6,  
				xlabel=Parallel Data, 
				ylabel=BPW, 
				grid=major, 
				grid style={dotted,black}, 
				style=ultra thick,
				xtick={0, 1, 2, 3, 4, 5}, 
				xticklabels = {0, 1K, 10K, 100K, 500K, 1M}
				]
				\pgfplotstableread[col sep = comma]{para_data_bpw.csv}\BPWScores
				\addplot[color=color1] table[x index = {0}, y index = {1}, mark=none]{\BPWScores};
			\end{axis}
			
			\begin{axis}[
				mark=none, 
				xmax=6,  
				ylabel=Sem-F1, 
				axis y line*=right, 
				axis x line=none,
				style=ultra thick,
				]
				\pgfplotstableread[col sep = comma]{para_data_srl_f1.csv}\FScores
				\addplot[color=color3] table[x index = {0}, y index = {1}, mark=none]{\FScores};
			\end{axis}
		\end{tikzpicture}
	}
	\caption{Language modeling effects vs. Parallel data size on the evaluation set.}\label{fig:para}
\end{figure}
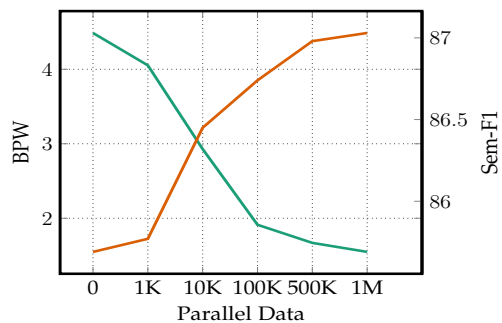

\begin{table*}[h]
	\small
	\begin{center}
		\small
		\setlength{\tabcolsep}{2pt}
		\caption{Comparison of the training and migration costs of the PrLMs.}
		\label{tab:cost}
		\begin{tabular}{l r l r r l r r}
			\toprule
			\textbf{Model} & \textbf{Data} & \textbf{BSZ} & \textbf{Steps}  & \textbf{Params} & \textbf{Hardware}  & \textbf{Train Time} & \textbf{G/TPU$\cdot$Days} \\
			\midrule
			ELMo & $\approx$4GB & $-$ & $-$ & 96M & 3 GTX 1080 GPUs & 14d & 42 \\
			GPT & $\approx$4.5GB &  64 & $\approx$1.2M  & 117M & 8 P6000 GPUs  & 25d & 200 \\
			BERT-base &  16GB & 256 & 1M & 110M & 16 TPUs / 8 V100 GPUs & 4d / 11d & 64 / 88 \\
			BERT-large & 16GB & 256 & 1M & 340M & 64 TPUs / 8 V100 GPUs & 4d  / 21d & 256 / 168  \\
			RoBERTa-large & 160GB & 8K & 500K & 340M & 1024 V100 GPUs & 1d  & 1024 \\
			XLNet-large & 126GB & 2K & 500K & 360M & 512 TPUs & 2.5d & 1280 \\
			ELECTRA-small & 16GB & 128 & 1M & 14M & 1 V100 GPU  & 4d  & 4 \\
			ELECTRA-base & 16GB & 256 & 766K & 110M & 64 TPUs & 4d  & 256 \\
			\midrule
			BERT-small &  $\approx$2.4GB  & 256  & 240K & 15M & 8 V100 GPUs & 1.5d  & 12 \\
			\bf TRI-BERT-base &\tikzmark{c}{$\approx$2.4GB} & 128 &  120K& 154M &  \tikzmark{a}{8 V100 GPUs} & 2.5d & 20 \\
			\bf TRI-BERT-large & &  128 & 120K & 398M & & 5d & 40 \\
			\bf TRI-RoBERTa-base & & 128   & 120K & 154M &  & 2.5d & 20\\
			\bf TRI-RoBERTa-large & \tikzmark{d}{$\approx$2.4GB} &  128 & 120K & 398M & \tikzmark{b}{8 V100 GPUs} & 5d & 40\\
			\bottomrule
		\end{tabular} 
		\link{a}{b}\link{c}{d}
	\end{center}
	\vspace{-1mm}
\end{table*}

\section{Discussion}

\subsection{Effects of Parallel Data Scale} 

Since the proposed \textsc{TreLM} framework relies on parallel corpora to learn the language differences explicitly, the sizes of the parallel corpora used are also of concern. We explored the influence of different parallel corpus sizes on the performance of the models transferred with the TRI-RoBERTa-base architecture.
The variation curve of BPW score with the size of parallel data is shown in Figure \ref{fig:para}. We see that with increasingly more parallel data, BPW gradually decreases, but this decrease slows as the data grows. The effect of the parallel corpora for cross-lingual transfer therefore has a upper bound because when the parallel corpora reaches a certain size, the errors from the alignment extraction tools cannot be ignored, and additionally, due to how lightweight the TRILayer structure is, TRILayers can only contain so much cross-lingual transfer information, which further restricts the growth of the migration performance.

\subsection{Pre-training Cost vs. Migration Training Cost} The training cost is an important factor for choosing whether to pre-train from scratch or to migrate from an existing PrLM. We listed the training data size, model parameters, training hardware, and training time of several public PrLM models and compared them with our models. The comparisons are shown in Table \ref{tab:cost}. Although the training hardware and engineering implementation of various PrLM models are different, this can still be used as a general reference. When model size is the same, our proposed model transferring framework is much faster than pre-training from scratch, and less data is used in the model transferring process. In addition, the total training time of our large model migration training is less than that of even the base model pre-training when hardware is kept the same. Therefore, the framework we proposed can be used as a good supplementary scheme for the PrLM in situations when time or computing resources are restricted.

\section{Ablations}\label{subsec:ablations}

\subsection{Effects of Different Embedding Initialization} 

To show the effectiveness of non-contextualized simulation and adversarial embedding space alignment, we compare the TRI-RoBERTa-base models obtained in the commonality training phase of our framework under four different embedding initialization configurations: \textit{random}, \textit{random+adversarial align}, \textit{fastText pre-trained}, and \textit{fastText pre-trained+adversarial align}. In addition, to lessen the influence of training different amounts during different initializations, we trained an additional 40K update steps in the commonality training phase. We selected newstest2020-enzh.ref.zh in the WMT-20 news translation task as the evaluation set with a total of 1418 sentences to avoid potential overlapping with the training set. The subword-level bits-per-word (BPW)  was used as the evaluation metric for the model's MLM performance\footnote{We do this because these models in comparison use the same vocabulary, and the masked parts on the evaluation set are identical, making the BPW scores comparable.}.

The BPW results on the evaluation set are presented in Table \ref{table:init}. The non-contextualized fastText embedding simulation and adversarial embedding alignment setting achieves better BPW scores than other configurations, which shows the effectiveness of our proposed approach. In addition, comparing the embedding initialization of \textit{random+adversarial align} and \textit{fastText pre-trained} shows that pre-training non-contextualized embeddings using language data is more effective than direct embedding space alignment. Considering different training 20K steps versus 40K steps, longer training leads to lower BPW, but the performance gains are less than what our method brings.

\begin{table}[h]
	\centering
	\caption{Evaluation of the subword-level BPW performance for the MLM objective of TRI-RoBERTa-base on various embedding initialization configurations after the commonality training.}\label{table:init}	
	\begin{tabular}{lcc}
		\toprule
		{\bf Configurations} & 20K & 40K  \\ 
		\midrule
		Rand & 6.9877 & 6.3264 \\ 
		Rand + Adv Align& 6.7725 & 5.9982 \\
		fastText PT & 6.5288 & 5.7957 \\
		fastText PT + Adv Align & 5.9330 & 5.2679 \\
		\bottomrule
	\end{tabular}
\end{table}

\begin{table}[h]
	\centering
	\setlength{\tabcolsep}{3pt}
	\footnotesize
	\caption{Evaluation of the translation performance of our migrated language models on the WMT \textit{newstest2020} test set with BLEU-1/2/3/4 metrics.}\label{table:nmt}
	\begin{tabular}{lcc}
		\toprule
		Models & EN$\rightarrow$ZH & ZH$\rightarrow$EN  \\ 
		\midrule
		Transformer-base  & 65.0 / 42.1 / 28.2 / 19.8 & 57.3 / 28.7 / 16.6 / 10.3 \\ 
		\midrule
		TRI-RoBERTa-base & 46.6 / 5.7 / 0.8 / 0.1 & 36.4 / 5.2 / 0.7 / 0.1 \\ 
		\quad w/o CdLM & 1.3 / 0.1 / 0.0 / 0.0 & 1.7 / 0.1 / 0.0 / 0.0 \\
		TRI-RoBERTa-large & 47.2 / 5.9 / 0.9 / 0.1 &  36.8 / 5.5 / 0.7 / 0.1 \\
		\quad w/o CdLM & 1.5 / 0.1 / 0.0 / 0.0 & 2.0 / 0.1 / 0.0 / 0.0 \\
		\bottomrule
	\end{tabular}
\end{table}

\subsection{Effects of Cross-lingual Model Transferring in \textsc{TreLM}} 

We conduct further ablation studies to analyze our proposed \textsc{TreLM} framework's cross-lingual model transferring design choices, including introducing the novel training objective, CdLM, and the TRILayer structure. The translation performance evaluation results are shown in Table \ref{table:nmt}. Using the newstest2020 en-zh and zh-en test sets, we evaluate the TRI-RoBERTa-base and TRI-RoBERTa-large models at the end of their transfer training phases. Since there is no alignment information available during the evaluation phase, we use the same successive alignment that MLM uses. For the sequence generated by the model, continuous repetitions were removed and the {\tt [SEP]} token was taken as the stop mark to obtain the final translation sequence. In the EN$\rightarrow$ZH translation direction, we report character-level BLEU, while in ZH$\rightarrow$EN, we report word-level BLEU. The Transformer-base NMT models for comparison are from \cite{TiedemannThottingal:EAMT2020} and were trained on the OPUS corpora \cite{tiedemann-2012-parallel}. 

As seen from the results, our TRI-RoBERTa-base and TRI-RoBERTa-large with CdLM were able to obtain very good BLEU-1 scores, indicating that the mapping between the transferring source language and target language was explicitly captured by the model. When CdLM is removed and we only use the traditional joint MLM and TLM for training on the same size parallel data, we find that the BLEU-1 score significantly decreases, demonstrating that joint MLM and TLM do not learn explicit alignment information. The BLEU-1 score is lower than that of the Transformer-base NMT model, but this is because the Transformer-base model uses more parallel corpora as well as a more complex model design compared to our non-autoregressive translation pattern and lightweight TRILayer structure. In addition, compared with BLEU-2/3/4, it can be seen that although Transformer-base can accurately translate some tokens, many tokens are not translated or are translated in the wrong order due to the lack of word ordering information and the differing sequence lengths, which result in a very low score. This also shows that word order is a very important factor in translation.

\begin{table}
	\centering
	\setlength{\tabcolsep}{3pt}
	\caption{\textcolor{black}{Language modeling effects of the CdLM objective and TRILayer structure for the Chinese TRI-RoBERTa-base model. UAS and LAS scores are given for the CTB 5.1 test set.}}
	\label{table:baseline}
	\begin{tabular}{lcccc} 
		\toprule
		Models   & Params & BPW & UAS & LAS \\ 
		\midrule
		TRI-RoBERTa-base & 154M & 1.548 &  90.30 & 87.82 \\
		\quad w/o CdLM & 154M &  3.028 &  88.16 & 85.20 \\
		\quad w/o CdLM \& w/o TRILayer & 148M & 3.469 & 87.45 & 84.69 \\
		\bottomrule
	\end{tabular}
\end{table}

Since the \textsc{TreLM} framework is evaluated using the existed pre-trained models, our migrated models are always larger than the original ones. Additional parameters arise in two places:  embedding layer parameters grow due to a larger vocabulary and language embeddings, and the TRILayer structure adds parameters. The embedding layer growth is necessary, but the TRILayer structure is optional, as it is only used for cross-lingual transfer training. Therefore, for this ablation, we test removing the TRILayer structure for a fairer comparison\footnote{In this setting, we train the model with same number of update steps using joint MLM and TLM when leveraging parallel sentences.} and show the results in Table \ref{table:baseline}.  Comparing the  evaluation set BPW scores of the final models obtained from RoBERTa-base under different migration methods, we found that our \textsc{TreLM} framework is stronger in cross-lingual model transferring compared to jointly using MLM and TLM, and it does not simply rely on the extra parameters of the TRILayer. \textcolor{black}{Furthermore, applying these pre-trained language models to the downstream task, dependency parsing on the CTB 5.1 treebank, achieves corresponding effects in BPW, which shows that the BPW score does describe the performance of PrLMs and that the pre-training performance will greatly affect performance in downstream tasks.}

\subsection{\textcolor{black}{Comparison of Different Cross-lingual Model Transferring Objectives}}

\textcolor{black}{As discussed in Section \ref{subsec:cdlm} , CdLM, TLM, and TLM variants such as BRLM are typical objectives of cross-lingual model transferring, in which parallel sentences are utilized for cross-lingual optimization. In order to compare the differences between these objectives empirically, we conducted a comparative experiment on TRI-RoBERTa-base. For this experiment, instead of using the model transferring objective CdLM in the second stage of training like our other models, we use TLM or BRLM instead. In addition, we follow \cite{artetxe-etal-2020-cross} in experimenting with the effects of joint vocabulary versus a separate vocabulary in cross-lingual model transferring, and we include a model, CdLM$^*$, with a separate vocabulary in this comparison as well. Specifically, for this model, we forego language embeddings and adopt independent token embeddings for difference languages. CdLM and MLM  alternately optimize the model.}

\textcolor{black}{The empirical comparison of these objectives is listed in Table 8. The migration target language is Chinese, and BPW score is used to compare the performance of the migrated model. We also show the dependency parsing performance on the CTB 5.1 dataset for the obtained model. Looking at CdLM and CdLM$^*$, in our \textsc{TreLM} framework, using a joint vocabulary leads to better performance than using a separate vocabulary strategy, which is not consistent with \cite{artetxe-etal-2020-cross} 's conclusion. We attribute this difference to the fact that \cite{artetxe-etal-2020-cross}'s model uses joint MLM pre-training of multiple languages to achieve implicit model transferring, so maintaining independent embeddings is important for distinguishing the language. In \textsc{TreLM}, because it trains two half-models, the explicit conversion signal guides the model's migration training in discerning the language. When using separate vocabularies, some common information (such as punctuation, loanwords, etc.) are ignored, lessening the impact of CdLM. Second, comparing TLM, BRLM, and CdLM, we note that CdLM takes the source and target language sequences as input and output, respectively, which cooperates with the TRILayer and half-model training strategy much better, whereas TL and BRLM combine the source and target sentences as input and predict a masked sentence as in MLM, which is much less conducive to the half-model training strategy. Because the source and target language sentences are separate in CdLM, the model is much more able to differentiate the two languages, which makes CdLM a stronger cross-lingual transferring objective.}

\begin{table}
	\centering
	\caption{\textcolor{black}{Effects of different cross-lingual transferring objectives. $^*$ indicates that a separate vocabulary is used.}}
	\label{table:different_objective}
	\begin{tabular}{lccc} 
		\toprule
		TRI-RoBERTa-base    & BPW & UAS & LAS \\ 
		\midrule
		w/ CdLM  & 1.548 & 90.30 & 87.82 \\
		w/ CdLM$^*$ & $-$ & 90.02 & 87.49 \\
		w/ TLM  & 3.028 &  88.16 & 85.20 \\
		w/ BRLM  & 2.932 & 89.85 & 87.27 \\
		\bottomrule
	\end{tabular}
\end{table}

\subsection{\textcolor{black}{Comparison with Cross-lingual Model Transferring Related Works on mPrLM}}

\begin{table}
	\centering
	\caption{\textcolor{black}{Performance of different cross-lingual model transferring approaches on dependency parsing on CTB 5.1.}}
	\label{table:related_work}
	\begin{tabular}{lcc} 
		\toprule
		Models & UAS & LAS \\ 
		\midrule
		m-BERT-base & 89.84 & 87.33 \\
		\hdashline
		+Target-Language Finetune & 90.28 & 87.76 \\
		+\textsc{RositaWORD} \cite{mulcaire-etal-2019-polyglot} & 89.88 & 87.36 \\
		+MIM \cite{liu-etal-2019-investigating} & 90.09 & 87.42 \\
		+Word-Alignment Finetune \cite{Cao2020Multilingual} & 90.33 & 87.79 \\
		\hdashline
		\bf TRI-BERT-base & 89.96 & 87.43 \\
		\bf TRI-BERT-base (400K) & 90.85 & 88.39  \\
		\bf TRI-m-BERT-base & 90.68 & 88.24 \\
		\bf TRI-m-BERT-base (400K) & 90.72 & 88.29 \\
		\bottomrule
	\end{tabular}
\end{table}

\textcolor{black}{Although we propose our method as an alternative to mPrLMs for cross-lingual transferring, it can also be applied to transfer the learning of mPrLMs. When transferring mPrLMs, the vocabulary replacement and embedding re-initialization are no longer needed, which makes our framework more simple.}

\textcolor{black}{We examine four main related approaches in the line of cross-lingual model transferring based on PrLMs. The first approach is trivial: using data from the target language and MLM to finetune a mPrLM. This helps specify the mPrLM as a PrLM specifically for the target language.}

\textcolor{black}{The second is \textsc{RositaWORD} \cite{mulcaire-etal-2019-polyglot}. In this method, the contextualized embeddings of mPrLM are concatenated with non-contextualized multilingual word embeddings. This representation is then aligned across languages in a supervisory manner using a parallel corpus, biasing the model toward cross-lingual feature sharing. The third, proposed by \cite{liu-etal-2019-investigating}, makes use of MIM (Meeting-In-the-Middle) \cite{doval-etal-2018-improving}, which uses a linear mapping to refine the embedding alignment, and is somewhat similar to our first step's adversarial embedding alignment, but because \cite{liu-etal-2019-investigating} only migrate the contextualized embedding of an mPrLM, it is not a true migration of the model. Specifically, their post-processing  trained linear mapping after the contextualized embedding of mPrLM is completely different from our new initialization of the raw embedding of PrLM. The fourth approach, Word-alignment Finetune, is similar in motivation to our CdLM, which uses the alignment information of the parallel corpora to perform finetuning training on the model (whereas \textsc{RositaWORD} and MIM focus on language-specific post-processing on the contextualized embedding of mPrLM). The difference is that Word-Alignment Finetune uses contextualized embedding similarity measurement for alignment to calculate the loss, and our method is inspired by machine translation, which uses language-to-language sequence translation for cross-lingual language modeling.}

\textcolor{black}{We evaluate the effectiveness of these methods on dependency parsing as shown in Table \ref{table:related_work}. We chose the widely used m-BERT-base as the base mPrLM and Chinese as the target language for these experiments. The resulting models were evaluated on the CTB 5.1 data of the dependency parsing task. For \textsc{RositaWORD}, we used the word-level embedding trained by Fastext and aligned by MUSE, as done in the original paper. For MIM, the number of training steps for the linear mapping is kept the same as in our first stage's adversarial embedding alignment training, and both train for 5 epochs. Target-Language Finetuning and Word-Alignment Finetuning use the same data as our main experiments and the same 120K update as well. We also listed a model migrated from a monolingual PrLM (TRI-BERT) to compare the performance differences between model transferring from monolinguals and multilingual PrLMs. Since the migrated mPrLM is simpler - it does not need to re-initialize or train embeddings and can converge faster, we train the migrated PrLM model longer steps (400K total training steps) to more fairly compare them. } 

\textcolor{black}{Comparing our \textsc{TreLM} with similar methods, the concatenation of cross-lingual aligned word-level embeddings in \textsc{RositaWORD} seems to have limited effect. MIM, which uses mapping for post-processing, leads to some improvement, but compared to Target-Language Finetune and Word-Alignment Finetune, it is obviously a weaker option. The results of TRI-m-BERT-base, Word-Alignment Finetune, and Target-Languagde Finetune suggest that using explicit alignment signals is advantageous compared to using the target language monolingual data when finetuning a limited amount of update steps, though when data is sufficient and training time is long enough, the performance for cross-lingually transferred models  will approach the performance of monolingually pre-trained models regardless of transfer method. Thus, the methods primarily differ in how they perform with limited data, computing resources, or time.
Our TRI-m-BERT-base outperforms +Word-Alignment Finetune, which shows that our CdLM, a language sequence modeling method inspired by machine translation, is more effective than solely deriving loss from an embedding space alignment.
The results of TRI-BERT-base and TRI-m-BERT-base demonstrate that the simpler migration for m-BERT-base provides an initial performance boost when both models are trained 120k steps due to its faster convergence, but when they are trained to a longer 400K steps, TRI-BERT-base actually shows better performance compared to TRI-m-BERT-base.}

\subsection{\textcolor{black}{More Languages for a More Comprehensive Evaluation}}

\textcolor{black}{In order to demonstrate the generalization ability of the cross-lingual model transferring of the proposed \textsc{TreLM} framework, we also migrate to German (DE) and Japanese (JA) in addition to Chinese and Indonesian. We also experimented with these languages on the Universal dependency parsing task.}

\textcolor{black}{The migrated German and Japanese TRI-BERT-base and TRI-RoBERTa-base use the same corpus size and training steps as their respective Chinese and Indonesian models. We show the results of German, and Japanese on UD in Table \ref{table:udp}. Since there are no official BERT-base models for these three language, we use third-party pre-trained models: Deepset BERT-base-german\footnote{\url{https://deepset.ai/german-bert}}, CL-TOHOKU BERT-base-japanese\footnote{\url{https://github.com/cl-tohoku/bert-japanese}}, and NICT BERT-base-japanese\footnote{\url{https://alaginrc.nict.go.jp/nict-bert/index.html}}.} 

\textcolor{black}{First, according to the results in the table, our TRI-BERT-base achieves quite similar performance compared to the third-party BERT-base models and  even exceeds the third-party models in some instances. This demonstrates that our \textsc{TreLM} is a general cross-lingual model transferring framework. Second, comparing third-party pre-trained BERT-base models and the official m-BERT-base, we found that some third-party BERTs are even less effective than m-BERT (Generally speaking, m-BERT is not as good as monolingual BERT when the data and training time are sufficient). This shows that in some scenarios, pre-training from scratch is not a very good choice, potentially due to insufficient data, unsatisfactory pre-training resource quality, and/or insufficient pre-training time. Compared with the well-trained monolingual BERT models, our migrated models are very competitive and can exceed PrLMs suffering from poor pre-training. In addition, in DE and JA, we also observed that the effect of TRI-RoBERTa was stronger than that of the TRI-BERT, indicating that our migration process maintained the performance advantage of the original model.}

\begin{table}
	\centering
	\setlength{\tabcolsep}{3pt}
	\caption{\textcolor{black}{Universal Dependency v2.3 parsing performance. $^*$ means that the results are evaluated based on our own implementation, not reported by \cite{dozat2016deep}. We use the following PrLMs not provided by the official \cite{devlin-etal-2019-bert}, third-party BERT-base PrLMs: Deepset BERT-base-german, CL-TOHOKU BERT-base-japanese ($\ddag$), and NICT BERT-base-japanese ($\S$).}}
	\label{table:udp}
	\begin{tabular}{lccccc} 
		\toprule
		\multirow{2}{*}{\bf Models} & \multicolumn{2}{c}{\bf DE GSD} &  & \multicolumn{2}{c}{\bf JA GSD} \\  
		\cmidrule{2-3} \cmidrule{5-6} & UAS & LAS &  & UAS & LAS \\ 
		\midrule
		\footnotesize(Dozat and Manning, 2016) \cite{dozat2016deep} & 86.84$^*$  &  81.31$^*$  &  & 86.24$^*$ & 84.52$^*$ \\  
		\multirow{2}{*}{BERT-base} & 89.28$\  $ & 84.60$\  $ & & 93.85$^\ddag$ & 91.62$^\ddag$ \\ 
		& &&&94.66$^\S$ & 92.65$^\S$\\ 
		m-BERT-base & 88.37$\  $ & 83.22$\  $ &  & 94.12$\  $ & 92.35$\  $ \\ 
		\hdashline
		\bf TRI-BERT-base & 89.12$\  $ & 84.34$\  $ &  & 94.30$\  $ & 92.56$\  $  \\ 
		\bf TRI-RoBERTa-base & 89.79$\  $ & 84.95$\  $ & & 94.98$\  $ & 93.01$\  $ \\ 
		\bottomrule
	\end{tabular}
\end{table}

\section{Conclusion and Future Work}
In this work, we present an effective method of transferring knowledge from a given language's pre-trained contextualized language model to a model in another language. This is an important accomplishment because it allows more languages to benefit from the massive improvements arising from these models, which have been primarily concentrated in English. As a further plus, this method also enables more efficient model training, as languages have commonalities, and models in the target language can exploit these commonalities and quickly adopt these common features rather than learning them from scratch. 
In future work, we plan to use our framework to transfer other models such as ELECTRA and models for more languages. We also aim to develop an unsupervised cross-lingual transferring objective to remove the reliance on parallel sentences.

\ifCLASSOPTIONcaptionsoff
\newpage
\fi

\bibliographystyle{IEEEtran}
\bibliography{references}

\appendices

\section{Analyzing Non-contextualized Embeddings and PrLMs' Raw Embeddings}\label{subsec:non_emb_simulation}

Bidirectional PrLMs such as BERT \cite{devlin-etal-2019-bert} use Masked Language Modeling (MLM) as the training objective, in which the model is required to predict a masked part of the sentence. This training paradigm has no essential difference with word2vec \cite{mikolov2013efficient}. Word2vec employed a simple single-layer perceptron neural network and restricted the context for the masked part to the sliding window, while recent mainstream PrLMs adopted self-attention-based Transformer as the context encoder, which can utilize the whole sentence as context. Because of this, we speculate that BERT's raw embeddings and word2vec embeddings have a similar nature, and that we can simulate BERT's raw embeddings with the word2vec embeddings through some special designs.

\begin{figure*}
	\begin{minipage}[t]{0.5\linewidth} 
		\centering 
		\begin{tikzpicture}
			\tikzstyle{every node}=[font=\Large,scale=0.45]
			\begin{axis}[
				xmin=0,xmax=4000,
				ymin=-0.15,ymax=0.4,
				ylabel=Cosine Value,
				xlabel=Quantity,
				enlargelimits=false,
				xbar interval,
				width=220pt,
				ytick={},
				y tick label style={/pgf/number format/.cd,%
					scaled y ticks = false,
					set thousands separator={},
					fixed},
				x tick label style={/pgf/number format/.cd,%
					scaled x ticks = false,
					set decimal separator={,},
					fixed}%
				]
				\addplot coordinates {
					(0, -1.00)
					(0, -0.98)
					(0, -0.96)
					(0, -0.94)
					(0, -0.92)
					(0, -0.90)
					(0, -0.88)
					(0, -0.86)
					(0, -0.84)
					(0, -0.82)
					(0, -0.80)
					(0, -0.78)
					(0, -0.76)
					(0, -0.74)
					(0, -0.72)
					(0, -0.70)
					(0, -0.68)
					(0, -0.66)
					(0, -0.64)
					(0, -0.62)
					(0, -0.60)
					(0, -0.58)
					(0, -0.56)
					(0, -0.54)
					(0, -0.52)
					(0, -0.50)
					(0, -0.48)
					(0, -0.46)
					(0, -0.44)
					(0, -0.42)
					(0, -0.40)
					(0, -0.38)
					(0, -0.36)
					(0, -0.34)
					(0, -0.32)
					(0, -0.30)
					(0, -0.28)
					(0, -0.26)
					(0, -0.24)
					(0, -0.22)
					(0, -0.20)
					(2, -0.18)
					(5, -0.16)
					(3, -0.14)
					(16, -0.12)
					(36, -0.10)
					(58, -0.08)
					(98, -0.06)
					(177, -0.04)
					(319, -0.02)
					(547, 0.00)
					(844, 0.02)
					(1375, 0.04)
					(2039, 0.06)
					(2746, 0.08)
					(3444, 0.10)
					(3919, 0.12)
					(3646, 0.14)
					(3073, 0.16)
					(2408, 0.18)
					(1762, 0.20)
					(1231, 0.22)
					(627, 0.24)
					(268, 0.26)
					(199, 0.28)
					(86, 0.30)
					(22, 0.32)
					(19, 0.34)
					(9, 0.36)
					(3, 0.38)
					(5, 0.40)
					(0, 0.42)
					(3, 0.44)
					(2, 0.46)
					(1, 0.48)
					(2, 0.50)
					(0, 0.52)
					(0, 0.54)
					(0, 0.56)
					(0, 0.58)
					(0, 0.60)
					(1, 0.62)
					(0, 0.64)
					(0, 0.66)
					(0, 0.68)
					(0, 0.70)
					(0, 0.72)
					(0, 0.74)
					(0, 0.76)
					(0, 0.78)
					(0, 0.80)
					(0, 0.82)
					(0, 0.84)
					(0, 0.86)
					(0, 0.88)
					(0, 0.90)
					(0, 0.92)
					(0, 0.94)
					(0, 0.96)
					(0, 0.98)
				};
			\end{axis}
		\end{tikzpicture}
	\end{minipage}%
	\begin{minipage}[t]{0.5\linewidth} 
		\centering 
		\begin{tikzpicture}
			\tikzstyle{every node}=[font=\Large,scale=0.45]
			\begin{axis}[
				xmin=0,xmax=300000,
				ymin=-0.15,ymax=0.3,
				xlabel=Quantity,
				ylabel=Cosine Value,
				enlargelimits=false,
				xbar interval,
				width=220pt,
				ytick={},
				y tick label style={/pgf/number format/.cd,%
					scaled y ticks = false,
					set thousands separator={},
					fixed},
				x tick label style={/pgf/number format/.cd,%
					scaled x ticks = false,
					set decimal separator={,},
					fixed}%
				]
				\addplot coordinates {
					(0, -1.00)
					(0, -0.98)
					(0, -0.96)
					(0, -0.94)
					(0, -0.92)
					(0, -0.90)
					(0, -0.88)
					(0, -0.86)
					(0, -0.84)
					(0, -0.82)
					(0, -0.80)
					(0, -0.78)
					(0, -0.76)
					(0, -0.74)
					(0, -0.72)
					(0, -0.70)
					(0, -0.68)
					(0, -0.66)
					(0, -0.64)
					(0, -0.62)
					(0, -0.60)
					(0, -0.58)
					(0, -0.56)
					(0, -0.54)
					(0, -0.52)
					(0, -0.50)
					(0, -0.48)
					(0, -0.46)
					(0, -0.44)
					(0, -0.42)
					(0, -0.40)
					(0, -0.38)
					(0, -0.36)
					(0, -0.34)
					(0, -0.32)
					(0, -0.30)
					(0, -0.28)
					(0, -0.26)
					(0, -0.24)
					(0, -0.22)
					(22, -0.20)
					(113, -0.18)
					(455, -0.16)
					(1584, -0.14)
					(4951, -0.12)
					(13208, -0.10)
					(30947, -0.08)
					(61553, -0.06)
					(109712, -0.04)
					(173643, -0.02)
					(237026, 0.00)
					(279301, 0.02)
					(284743, 0.04)
					(251712, 0.06)
					(196316, 0.08)
					(136894, 0.10)
					(87860, 0.12)
					(54028, 0.14)
					(31549, 0.16)
					(18707, 0.18)
					(10866, 0.20)
					(6460, 0.22)
					(3840, 0.24)
					(2030, 0.26)
					(1130, 0.28)
					(631, 0.30)
					(339, 0.32)
					(174, 0.34)
					(83, 0.36)
					(57, 0.38)
					(27, 0.40)
					(9, 0.42)
					(11, 0.44)
					(3, 0.46)
					(3, 0.48)
					(2, 0.50)
					(1, 0.52)
					(3, 0.54)
					(1, 0.56)
					(1, 0.58)
					(1, 0.60)
					(0, 0.62)
					(0, 0.64)
					(1, 0.66)
					(0, 0.68)
					(2, 0.70)
					(0, 0.72)
					(0, 0.74)
					(0, 0.76)
					(0, 0.78)
					(0, 0.80)
					(0, 0.82)
					(0, 0.84)
					(0, 0.86)
					(0, 0.88)
					(0, 0.90)
					(0, 0.92)
					(0, 0.94)
					(0, 0.96)
					(0, 0.98)
				};
			\end{axis}
		\end{tikzpicture}
	\end{minipage} 
	\caption{Individual histogram plots of cosine similarity of a single term ``genes" with other terms in the vocabularies of BERT-base-cased (left) and FastText cc.en.300d (right).}
	\label{fig:emb_dist}
\end{figure*}
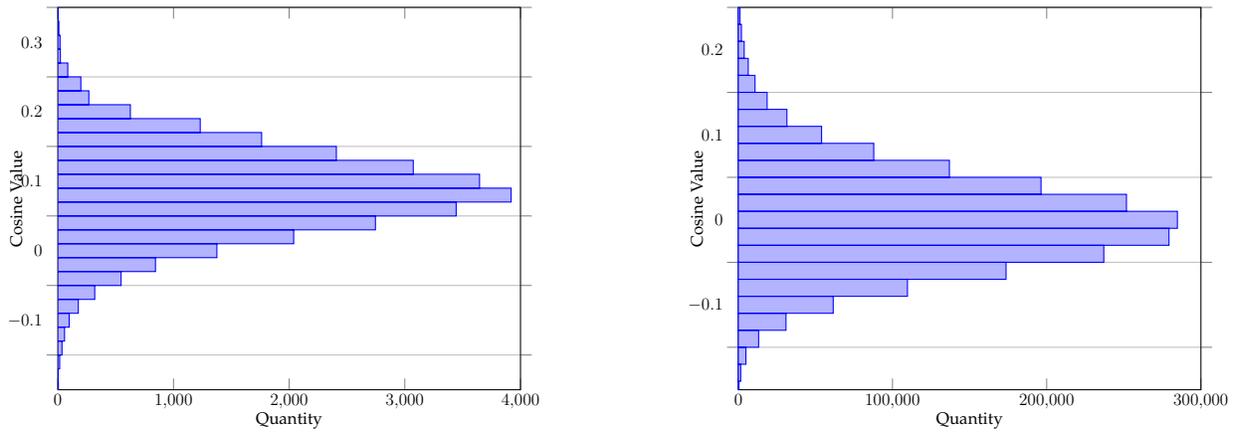

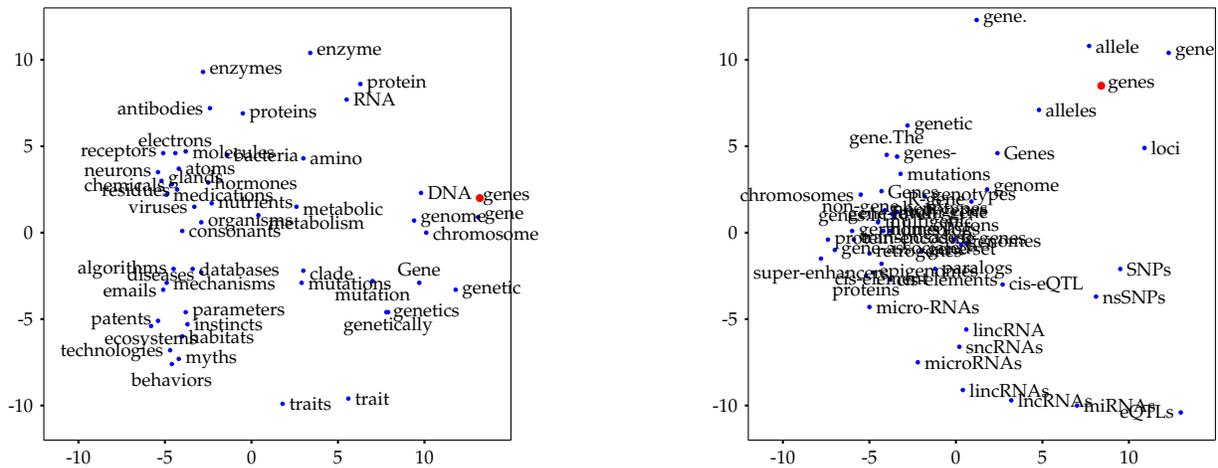
\begin{figure*}[h]
	\begin{minipage}[t]{0.5\linewidth} 
		\centering
		\scriptsize
		\begin{tikzpicture} [xscale=0.23, yscale=0.23]
			\draw [-] (-12,13) -- (-12,-12) -- (15,-12);
			\draw [-] (-12,13) -- (15,13) -- (15,-12);
			\draw [thick] (-10,-12.2) node[below]{-10} -- (-10,-12);
			\draw [thick] (-5,-12.2) node[below]{-5} -- (-5,-12);
			\draw [thick] (0,-12.2) node[below]{0} -- (0,-12);
			\draw [thick] (5,-12.2) node[below]{5} -- (5,-12);
			\draw [thick] (10,-12.2) node[below]{10} -- (10,-12);
			\draw [thick] (-12.2,-10) node[left]{-10} -- (-12,-10);
			\draw [thick] (-12.2,-5) node[left]{-5} -- (-12,-5);
			\draw [thick] (-12.2,0) node[left]{0} -- (-12,0);
			\draw [thick] (-12.2,5) node[left]{5} -- (-12,5);
			\draw [thick] (-12.2,10) node[left]{10} -- (-12,10);
			\draw [fill,red] (13.2, 2.0) circle [radius=0.2];
			\draw [fill,blue] (13.1, 0.9) circle [radius=0.1];
			\draw [fill,blue] (-0.5, 6.9) circle [radius=0.1];
			\draw [fill,blue] (7.8, -4.6) circle [radius=0.1];
			\draw [fill,blue] (9.4, 0.7) circle [radius=0.1];
			\draw [fill,blue] (11.8, -3.3) circle [radius=0.1];
			\draw [fill,blue] (2.9, -2.9) circle [radius=0.1];
			\draw [fill,blue] (-2.8, 9.3) circle [radius=0.1];
			\draw [fill,blue] (-2.5, 2.9) circle [radius=0.1];
			\draw [fill,blue] (1.8, -9.9) circle [radius=0.1];
			\draw [fill,blue] (10.1, 0.0) circle [radius=0.1];
			\draw [fill,blue] (-2.3, 1.7) circle [radius=0.1];
			\draw [fill,blue] (9.8, 2.3) circle [radius=0.1];
			\draw [fill,blue] (-5.1, 4.6) circle [radius=0.1];
			\draw [fill,blue] (-4.5, -2.1) circle [radius=0.1];
			\draw [fill,blue] (7.9, -4.6) circle [radius=0.1];
			\draw [fill,blue] (7.0, -2.8) circle [radius=0.1];
			\draw [fill,blue] (-5.4, 3.5) circle [radius=0.1];
			\draw [fill,blue] (-1.4, 4.5) circle [radius=0.1];
			\draw [fill,blue] (-3.7, -5.3) circle [radius=0.1];
			\draw [fill,blue] (-3.3, 1.5) circle [radius=0.1];
			\draw [fill,blue] (-4.9, 2.2) circle [radius=0.1];
			\draw [fill,blue] (6.3, 8.6) circle [radius=0.1];
			\draw [fill,blue] (-5.2, 3.0) circle [radius=0.1];
			\draw [fill,blue] (-4.2, 3.7) circle [radius=0.1];
			\draw [fill,blue] (-3.8, 4.7) circle [radius=0.1];
			\draw [fill,blue] (-4.6, -7.6) circle [radius=0.1];
			\draw [fill,blue] (-4.3, 2.5) circle [radius=0.1];
			\draw [fill,blue] (-5.4, -5.1) circle [radius=0.1];
			\draw [fill,blue] (5.6, -9.6) circle [radius=0.1];
			\draw [fill,blue] (-4.6, 2.8) circle [radius=0.1];
			\draw [fill,blue] (5.5, 7.7) circle [radius=0.1];
			\draw [fill,blue] (-2.9, 0.6) circle [radius=0.1];
			\draw [fill,blue] (3.4, 10.4) circle [radius=0.1];
			\draw [fill,blue] (-4.4, 4.6) circle [radius=0.1];
			\draw [fill,blue] (-2.4, 7.2) circle [radius=0.1];
			\draw [fill,blue] (-4.0, -6.0) circle [radius=0.1];
			\draw [fill,blue] (0.4, 1.0) circle [radius=0.1];
			\draw [fill,blue] (-5.8, -5.4) circle [radius=0.1];
			\draw [fill,blue] (-3.4, -2.1) circle [radius=0.1];
			\draw [fill,blue] (9.7, -2.9) circle [radius=0.1];
			\draw [fill,blue] (3.0, -2.2) circle [radius=0.1];
			\draw [fill,blue] (3.0, 4.3) circle [radius=0.1];
			\draw [fill,blue] (-4.0, 0.1) circle [radius=0.1];
			\draw [fill,blue] (-2.9, -2.3) circle [radius=0.1];
			\draw [fill,blue] (2.6, 1.5) circle [radius=0.1];
			\draw [fill,blue] (-4.7, -6.8) circle [radius=0.1];
			\draw [fill,blue] (-3.8, -4.6) circle [radius=0.1];
			\draw [fill,blue] (-5.1, -3.3) circle [radius=0.1];
			\draw [fill,blue] (-4.9, -2.9) circle [radius=0.1];
			\draw [fill,blue] (-4.2, -7.3) circle [radius=0.1];
			\node [right] at (13.0, 2.0) {genes};
			\node [right] at (13.1, 0.9) {gene};
			\node [right] at (-0.5, 6.9) {proteins};
			\node [right] at (7.8, -4.6) {genetics};
			\node [right] at (9.4, 0.7) {genome};
			\node [right] at (11.8, -3.3) {genetic};
			\node [above, right] at (2.9, -2.9) {mutations};
			\node [right] at (-2.8, 9.3) {enzymes};
			\node [right] at (-2.5, 2.9) {hormones};
			\node [right] at (1.8, -9.9) {traits};
			\node [right] at (10.1, -0.0) {chromosome};
			\node [right] at (-2.3, 1.8) {nutrients};
			\node [right] at (9.8, 2.3) {DNA};
			\node [above, left] at (-5.1, 4.6) {receptors};
			\node [left] at (-4.5, -2.1) {algorithms};
			\node [below] at (7.9, -4.5) {genetically};
			\node [below] at (7.0, -2.8) {mutation};
			\node [left] at (-5.4, 3.5) {neurons};
			\node [below, right] at (-1.4, 4.5) {bacteria};
			\node [right] at (-3.7, -5.3) {instincts};
			\node [left] at (-3.3, 1.5) {viruses};
			\node [right] at (-4.9, 2.2) {medications};
			\node [right] at (6.3, 8.6) {protein};
			\node [right] at (-5.2, 3.1) {glands};
			\node [above, right] at (-4.2, 3.7) {atoms};
			\node [right] at (-3.8, 4.7) {molecules};
			\node [below] at (-4.6, -7.7) {behaviors};
			\node [below, left] at (-4.3, 2.5) {residues};
			\node [left] at (-5.4, -5.1) {patents};
			\node [right] at (5.6, -9.6) {trait};
			\node [above, left] at (-4.6, 2.8) {chemicals};
			\node [right] at (5.5, 7.7) {RNA};
			\node [right] at (-2.9, 0.6) {organisms};
			\node [right] at (3.4, 10.4) {enzyme};
			\node [left, above] at (-4.4, 4.6) {electrons};
			\node [above, left] at (-2.4, 7.2) {antibodies};
			\node [right] at (-4.0, -6.0) {habitats};
			\node [below, right] at (0.6, 0.6) {metabolism};
			\node [left, below] at (-5.8, -5.4) {ecosystems};
			\node [above, right] at (-3.4, -2.1) {databases};
			\node [above] at (9.7, -2.9) {Gene};
			\node [above, right] at (3.0, -2.3) {clade};
			\node [right] at (3.0, 4.3) {amino};
			\node [right] at (-4.0, 0.1) {consonants};
			\node [left] at (-2.9, -2.4) {diseases};
			\node [above, right] at (2.6, 1.4) {metabolic};
			\node [left] at (-4.7, -6.9) {technologies};
			\node [right] at (-3.8, -4.6) {parameters};
			\node [left] at (-5.1, -3.3) {emails};
			\node [below, right] at (-4.9, -2.9) {mechanisms};
			\node [below, right] at (-4.2, -7.3) {myths};
		\end{tikzpicture}
	\end{minipage}
	\begin{minipage}[t]{0.5\linewidth} 
		\centering
		\scriptsize
		\begin{tikzpicture} [xscale=0.23, yscale=0.23]
			\draw [-] (-12,13) -- (-12,-12) -- (15,-12);
			\draw [-] (-12,13) -- (15,13) -- (15,-12);
			\draw [thick] (-10,-12.2) node[below]{-10} -- (-10,-12);
			\draw [thick] (-5,-12.2) node[below]{-5} -- (-5,-12);
			\draw [thick] (0,-12.2) node[below]{0} -- (0,-12);
			\draw [thick] (5,-12.2) node[below]{5} -- (5,-12);
			\draw [thick] (10,-12.2) node[below]{10} -- (10,-12);
			\draw [thick] (-12.2,-10) node[left]{-10} -- (-12,-10);
			\draw [thick] (-12.2,-5) node[left]{-5} -- (-12,-5);
			\draw [thick] (-12.2,0) node[left]{0} -- (-12,0);
			\draw [thick] (-12.2,5) node[left]{5} -- (-12,5);
			\draw [thick] (-12.2,10) node[left]{10} -- (-12,10);
			\draw [fill,red] (8.4, 8.5) circle [radius=0.2];
			\draw [fill,blue] (12.3, 10.4) circle [radius=0.1];
			\draw [fill,blue] (-4.3, 2.4) circle [radius=0.1];
			\draw [fill,blue] (4.8, 7.1) circle [radius=0.1];
			\draw [fill,blue] (-3.4, 4.4) circle [radius=0.1];
			\draw [fill,blue] (-2.9, 1.3) circle [radius=0.1];
			\draw [fill,blue] (-3.6, 0.9) circle [radius=0.1];
			\draw [fill,blue] (2.4, 4.6) circle [radius=0.1];
			\draw [fill,blue] (-3.2, 3.4) circle [radius=0.1];
			\draw [fill,blue] (9.5, -2.1) circle [radius=0.1];
			\draw [fill,blue] (-5.0, -1.2) circle [radius=0.1];
			\draw [fill,blue] (-0.1, -0.4) circle [radius=0.1];
			\draw [fill,blue] (0.3, -0.7) circle [radius=0.1];
			\draw [fill,blue] (-3.8, -2.7) circle [radius=0.1];
			\draw [fill,blue] (-4.1, 1.3) circle [radius=0.1];
			\draw [fill,blue] (0.4, -9.1) circle [radius=0.1];
			\draw [fill,blue] (0.2, -6.6) circle [radius=0.1];
			\draw [fill,blue] (10.9, 4.9) circle [radius=0.1];
			\draw [fill,blue] (1.8, 2.5) circle [radius=0.1];
			\draw [fill,blue] (-4.8, -2.3) circle [radius=0.1];
			\draw [fill,blue] (-4.0, 4.5) circle [radius=0.1];
			\draw [fill,blue] (-5.0, -4.3) circle [radius=0.1];
			\draw [fill,blue] (3.2, -9.7) circle [radius=0.1];
			\draw [fill,blue] (-1.2, -2.1) circle [radius=0.1];
			\draw [fill,blue] (2.7, -3.0) circle [radius=0.1];
			\draw [fill,blue] (0.6, -5.6) circle [radius=0.1];
			\draw [fill,blue] (-6.0, 0.1) circle [radius=0.1];
			\draw [fill,blue] (-3.4, 1.2) circle [radius=0.1];
			\draw [fill,blue] (-5.2, -2.5) circle [radius=0.1];
			\draw [fill,blue] (-4.3, -1.8) circle [radius=0.1];
			\draw [fill,blue] (-1.8, 2.1) circle [radius=0.1];
			\draw [fill,blue] (-5.5, 2.2) circle [radius=0.1];
			\draw [fill,blue] (0.9, 1.8) circle [radius=0.1];
			\draw [fill,blue] (-2.2, -7.5) circle [radius=0.1];
			\draw [fill,blue] (-7.0, -1.0) circle [radius=0.1];
			\draw [fill,blue] (7.0, -10.0) circle [radius=0.1];
			\draw [fill,blue] (7.7, 10.8) circle [radius=0.1];
			\draw [fill,blue] (-2.8, 6.2) circle [radius=0.1];
			\draw [fill,blue] (-2.0, -1.1) circle [radius=0.1];
			\draw [fill,blue] (-7.4, -0.4) circle [radius=0.1];
			\draw [fill,blue] (-3.8, 0.1) circle [radius=0.1];
			\draw [fill,blue] (8.1, -3.7) circle [radius=0.1];
			\draw [fill,blue] (-4.5, 0.6) circle [radius=0.1];
			\draw [fill,blue] (0.6, -0.6) circle [radius=0.1];
			\draw [fill,blue] (13.0, -10.4) circle [radius=0.1];
			\draw [fill,blue] (1.2, 12.3) circle [radius=0.1];
			\draw [fill,blue] (-4.2, 0.1) circle [radius=0.1];
			\draw [fill,blue] (-7.8, -1.5) circle [radius=0.1];
			\draw [fill,blue] (-5.9, -0.4) circle [radius=0.1];
			\draw [fill,blue] (-3.7, 1.1) circle [radius=0.1];
			\node [right] at (8.4, 8.5) {genes};
			\node [right] at (12.3, 10.4) {gene};
			\node [right] at (-4.3, 2.4) {Genes};
			\node [right] at (4.8, 7.1) {alleles};
			\node [right] at (-3.4, 4.4) {genes-};
			\node [left] at (-2.9, 1.3) {non-gene};
			\node [left] at (-3.6, 0.9) {genes.In};
			\node [right] at (2.4, 4.6) {Genes};
			\node [right] at (-3.2, 3.4) {mutations};
			\node [right] at (9.5, -2.1) {SNPs};
			\node [right] at (-5.0, -1.2) {retrogenes};
			\node [right] at (-0.1, -0.4) {R-genes};
			\node [right] at (0.3, -0.7) {genomes};
			\node [right] at (-3.8, -2.7) {cis-elements};
			\node [right] at (-4.1, 1.3) {phenotypes};
			\node [right] at (0.4, -9.1) {lincRNAs};
			\node [right] at (0.2, -6.6) {sncRNAs};
			\node [right] at (10.9, 4.9) {loci};
			\node [right] at (1.8, 2.5) {genome};
			\node [right] at (-4.8, -2.3) {epigenomes};
			\node [above] at (-4.0, 4.5) {gene.The};
			\node [right] at (-5.0, -4.3) {micro-RNAs};
			\node [right] at (3.2, -9.7) {lncRNAs};
			\node [right] at (-1.2, -2.1) {paralogs};
			\node [right] at (2.7, -3.0) {cis-eQTL};
			\node [right] at (0.6, -5.6) {lincRNA};
			\node [right] at (-6.0, 0.1) {germlines};
			\node [right] at (-3.4, 1.2) {polygenes};
			\node [left, below] at (-5.2, -2.5) {proteins};
			\node [below] at (-4.3, -1.8) {cis-element};
			\node [right] at (-1.8, 2.1) {genotypes};
			\node [left] at (-5.5, 2.2) {chromosomes};
			\node [left] at (0.9, 1.8) {R-gene};
			\node [right] at (-2.2, -7.5) {microRNAs};
			\node [right] at (-7.0, -1.0) {gene-associated};
			\node [right] at (7.0, -10.0) {miRNAs};
			\node [right] at (7.7, 10.8) {allele};
			\node [right] at (-2.8, 6.2) {genetic};
			\node [right] at (-2.0, -1.1) {gene-set};
			\node [right] at (-7.4, -0.4) {protein-encoding};
			\node [above] at (-3.8, 0.1) {gene-level};
			\node [right] at (8.1, -3.7) {nsSNPs};
			\node [right] at (-4.5, 0.6) {multigenic};
			\node [above] at (0.6, -0.6) {operons};
			\node [left] at (13.0, -10.4) {eQTLs};
			\node [right] at (1.2, 12.3) {gene.};
			\node [right] at (-4.2, 0.1) {homeologs};
			\node [below] at (-7.8, -1.5) {super-enhancers};
			\node [right] at (-5.9, -0.4) {transposases};
			\node [right] at (-3.7, 1.1) {multi-gene};
		\end{tikzpicture}
	\end{minipage}
	\caption{Two-dimensional PCA projection of the embedding vectors representing ``genes" and top-50 similar terms in vocabularies of BERT-base-cased and FastText cc.en.300d.}\label{fig:dist}
\end{figure*}

To verify our theory, we studied the important relational nature of embeddings. Specifically, we chose BERT-base-cased's raw embeddings and word2vec-based FastText cc.en.300d embeddings \cite{grave-etal-2018-learning} and evaluated the cosine similarity of single terms compared to other terms in their vocabularies. An example histogram for the term ``genes" is shown in Figure \ref{fig:emb_dist}. Examining the two types of embeddings, we found that  the learned vectors, regardless of the type of similarity (semantic/syntactic/inflections/spelling/etc.) they capture, have a very similar distribution shape. This showed us that the two embedding spaces are similar, and words within them may just have different relations to each other. Thus, our work focuses on aligning the new word2vec embedding space by learning a mapping to the original embedding space to simulate the original embedding allow for a cross-lingual migration of the PrLM. 

To illustrate the necessity of embedding alignment, we also took out the top-50 terms closest to the term ``genes" in the two embedding spaces, used principal component analysis (PCA) to reduce the vector dimension to 2, and presented it in a two-dimensional figure, as shown in Figure \ref{fig:dist}. As can be seen from the figure, due to the different language modeling architectures and contexts in FastText and BERT, corresponding points are distributed at different locations in the embedding space. This is why compatibility problems exist when we use the original non-contextualized embeddings to simulate the new embedding and hence why we need to align the embeddings. 

\section{Downstream Tasks}\label{subsec:tasks}

Following previous contextualized language model pre-training, we evaluated the English-to-Chinese migrated language models on the CLUE benchmark. The Chinese Language Understanding Evaluation (CLUE) benchmark \cite{xu2020clue} consists of six different natural language understanding tasks: Ant Financial Question Matching (AFQMC), TouTiao Text Classification for News Titles (TNEWS), IFLYTEK \cite{co2019iflytek}, Chinese-translated Multi-Genre Natural Language Inference (CMNLI),  Chinese Winograd Schema Challenge (WSC), and Chinese Scientific Literature (CSL) and three machine reading comprehension tasks: Chinese Machine Reading Comprehension (CMRC) 2018 \cite{cui-etal-2019-span},  Chinese IDiom cloze test (CHID) \cite{zheng-etal-2019-chid}, and Chinese multiple-Choice machine reading Comprehension (C$^3$) \cite{sun2019probing}. We built baselines for the natural language understanding tasks by adding a linear classifier on top of the ``{\tt [CLS]}" token to predict label probabilities. For the extractive question answering task, CMRC, we packed the question and passage tokens together with special tokens to form the input: ``{\tt [CLS]} $Question$ {\tt [SEP]} $Passage$ {\tt [SEP]}", and employed two linear output layers to predict the probability of each token being the start and end positions of the answer span following the practice for BERT \cite{devlin-etal-2019-bert}. Finally, in the multi-choice reading comprehension tasks, CHILD and C$^3$, we concatenated the passage, question, and each candidate answer (``{\tt [CLS]} $Question$ $||$ $Answer$ {\tt [SEP]} $Passage$ {\tt [SEP]}"), input this to the models, and also predicted the probability of each answer on the representations from the ``{\tt [CLS]}" token following prior works \cite{yang2019xlnet, liu2019roberta}.

In addition to these language understanding tasks, language structure analysis tasks are also a very important part of natural language processing. Therefore, we also evaluated the PrLMs on syntactic dependency parsing and semantic role labeling, a type of semantic parsing. The baselines we selected for dependency parsing and semantic role labeling are from \cite{dozat2016deep} and \cite{cai-etal-2018-full}, respectively. These two baseline models are very strong and efficient and rely only on pure model structures to obtain advanced parsing performance. Our approach to integrate the PrLM with the two baselines is to replace the BiLSTM encoder in the baseline with the encoder of the PrLM. We took the first subword or character representation of a word as the representation of a word, which solved the PrLM's inconsistent granularity issue that impeded parsing. 

For the English-to-Indonesian migrated language models, since the language understanding tasks in Indonesian are very limited, we chose to use the Universal Dependency (UD) parsing task (v2.3, \cite{zeman-etal-2018-conll}), in which the treebanks of the world's languages were built by an international cooperative project, as the downstream task for evaluation.

\end{document}